# A Gated Catchment Hydrology Model (GCHM) for Physically-Interpretable Rainfall-Evapotranspiration-Runoff Simulation Across Diverse Hydroclimatic Regimes

M Jawad[1], HV Gupta[1], MA Farmani[1], A Behrangi[1], and GY Niu[1]

[1]Department of Hydrology and Atmospheric Sciences, The University of Arizona, Tucson, AZ, USA, 85721.

## Key Points:

- A state-of-the-art, parsimonious, interpretable five-state gated architecture explicitly represents water storage, fluxes, and runoff pathways while learning nonlinear process relationships.
- GCHM reproduces streamflow dynamics accurately across rainfall-, snowmelt-, historical-rainfall-, and arid catchments, with systematic ablations revealing the value of key process representations.
- Its gates reveal plausible snowmelt, infiltration, seepage, and groundwater behaviors, while more flexible ET formulations improve robustness across contrasting regimes.

## Abstract:

Recent advances in machine learning have substantially improved rainfall–runoff prediction, yet most data-driven models remain difficult to interpret and do not explicitly enforce physical conservation laws. We present the *Gated Catchment Hydrology Model* (GCHM) model, a physically-interpretable, mass-conserving hydrologic architecture based on the Mass-Conserving Perceptron (MCP). The model represents major catchment processes through gated storage units that explicitly conserve mass while learning context-dependent process relationships directly from data. The five-state architecture includes canopy interception, snow accumulation and melt, upper soil moisture, groundwater storage, and streamflow generation, with physically constrained gating functions controlling precipitation partitioning, evapotranspiration, infiltration, seepage, runoff production, and baseflow.

The model was evaluated using 50 representative CAMELS catchments spanning *snowmelt*-dominated, *recent-rainfall*, *historical-rainfall*, and *arid* hydroclimatic regimes. Meteorological forcing was derived from NOAA AORC together with remotely sensed vegetation and land-surface datasets. Performance was assessed using the *Jawad-Kling-Gupta Efficiency* (JKGE) metric, which explicitly accounts for both stationary and non-stationary aspects of hydrologic behavior. We further compared the full architecture with systematically ablated model variants and recurrent neural network benchmarks.

Results demonstrate that the proposed architecture reproduces streamflow dynamics across diverse hydroclimatic conditions, including *arid* regimes, while maintaining physically-meaningful states and fluxes. Ablation experiments reveal that while simplified model structures may perform adequately under high-flow (wet) conditions, their deficiencies become evident under dry conditions (particularly in representing flow persistence and recession dynamics) where groundwater storage is important for sustaining low flows. Learned gates recover hydrologically-plausible relationships governing snowmelt, runoff, seepage, and evapotranspiration. Compared with simpler conceptual representations and pure-data-based networks, GCHM offers robust, interpretable and competitive predictive performance, demonstrating the potential of conservation-based machine learning to facilitate interpretable hydrologic modeling that can support counterfactual analysis.

## 1. Introduction, Background and Scope

### 1.1. Statement of the Problem

[1] Data-driven machine learning (ML) and deep learning (DL) methods have emerged as powerful tools for modeling catchment-scale rainfall-evapotranspiration-runoff (RER) dynamics. Their ability to flexibly learn complex input-output relationships directly from data, enables them to replicate nonlinear time-dependent behaviors without prescribing the detailed mathematical forms of every relevant hydro-meteorological process. Consequently, they demonstrate considerable power for streamflow simulation and other hydro-meteorological prediction tasks (*Sorooshian, 1983; Hsu et al, 1995; Kratzert et al., 2018; Nearing et al., 2024*). However, this same ability introduces an important challenge: the internal representations learned by conventional ML models do not necessarily correspond to human-recognizable hydro-meteorological states, fluxes, or processes, making it difficult to determine how incoming mass and energy drivers are partitioned, how water and heat are stored and released, or how individual flux pathways contribute to the simulated hydrometeorological response.

[2] So, while ML-based models may reproduce observed catchment behaviors effectively, they may provide limited information about the mechanisms responsible for those behaviors. This distinction is important when the scientific objective extends to an understanding of catchment functioning, and the consequent ability to conduct counterfactual analyses (e.g., how the system might behave under climatically or anthropogenically altered conditions).

[3] In this regard, traditional physical-conceptual (PC) models explicitly represent hydro-meteorological processes and relationships based on scientific understanding (*Zhao, 1992; Lindström et al., 1997; Perrin et al., 2003; Anderson et al., 2006*). Because their structures are explicitly based on human-interpretable representations of mass and energy stores, fluxes and process relationships, they allow assumptions to be directly examined and scientific hypotheses to be explored and tested, thereby providing a conceptual basis for interpreting model behavior. In contrast, typical ML architectures employ generic computational units whose internal states and connections have no prescribed correspondence with physical flux processes (*Kratzert et al., 2018; Kratzert et al., 2019*), which makes it difficult (if not impossible) to assess how hydro-meteorological inputs are partitioned, how mass and energy are stored and released, or how individual flux pathways contribute to the simulated hydro-meteorological response.

### 1.2. Bridging the Gap

[4] This gap between the predictive power of ML-based models and the human-interpretability (and support for counterfactual analysis) provided by PC models has driven interest in the development of approaches that can retain the predictive power and flexible-learning ability of data-driven methods while providing greater physical interpretability (*Rudin, 2019*), and several approaches have been proposed to bridge the gap. These include methods for (i) interpreting trained ML models, (ii) incorporating physical constraints, and (iii) integrating prior scientific knowledge into the learning process (*Althoff et al., 2021; Jiang et al., 2024*).

[5] For example, *Knowledge-Guided Machine Learning* (KGML) provides a general framework for incorporating domain knowledge into ML-based models, thereby reducing the disconnect between data-driven learning and established scientific understanding (*Willard et al., 2022; Varadharajan et al., 2022*). However, although KGML approaches can improve the physical consistency or interpretability of ML-based predictions, incorporating physical information into a model does not necessarily ensure that its internal computational structure has a direct and interpretable relationship with the underlying hydrological system. To facilitate advancements in process understanding, it is therefore desirable to require interpretability at the level of model architectural components themselves, rather than treating architecture solely as a property that is inferred after a generic neural network has been trained

[6] In this regard, model complexity presents an important challenge. By allowing large numbers of computational units and connections, ML-based models gain the flexibility to represent complex relationships, while simultaneously making it increasingly difficult to identify the essential mechanisms represented by the

model. From the perspective of human-interpretability, a useful scientific model is one that that (i) makes the relationship between system structure and system behavior explicit, and (ii) is representationally simple enough to facilitate understanding/reasoning (*Weijs & Ruddell, 2020*), while (iii) remaining complex enough to reproduce its important behaviors.

[7] One approach to ML-model simplification is to apply (after training) pruning or compression techniques that reduce network size while attempting to retain predictive skill (*Tishby & Zaslavsky, 2015; Shwartz-Ziv & Tishby, 2017; Schotthöfer et al., 2022*). Another is to start from a simple initial structure and progressively increase model complexity until the desired predictive power is achieved *(Hsu et al., 1995; Chen & Chang, 2009; Lan et al., 2022*). However, reducing unnecessary complexity does not necessarily achieve a physically meaningful representation that facilitates interpretation of the individual computational elements.

[8] A more direct approach is to explicitly require domain-relevant interpretability of the individual computational units (functional basis units) that are used to construct the model. In hydro-meteorological modeling, this provides the opportunity to design ML architectures in which the computational states and fluxes are explicitly (conceptually) related to the storage and movement of mass and energy, while allowing the behavioral forms of the relationships governing such storage and movement to be directly learned from data. A foundation for such an approach is provided by the *Mass-Conserving Perceptron* (MCP) introduced by *Wang & Gupta* (*2024a*), which formulates the computational unit around a discrete-time (extensible to continuous-time) mass-balance relationship while retaining the flexible-learning capability of neural networks. Because the MCP explicitly imposes conservation of the quantities entering, stored within, and leaving a computational unit, it provides a natural building block for developing ML-based representations of dynamical geoscientific systems. Subsequent applications demonstrated that MCP-based architectures can be used to construct interpretable catchment-scale hydrological models and investigate parsimonious representations of RER dynamics *(Wang & Gupta, 2024b, 2025*).

## 1.3. Scientific Contributions and Scope of this Study

### 1.3.1. The Gated Catchment Hydrology Model

[9] Building upon the MCP framework, the present study develops a *Gated Catchment Hydrology Model* (GCHM) architecture in which mass-conserving computational units are carefully selected and arranged to represent our current understanding of the major components and dominant processes that make up the catchment water cycle. Rather than treating the neural network as a generic mapping from hydro-meteorological inputs to responses, the GCHM explicitly represents a physical-conceptual sequence of interacting stores and pathways. The current version focuses on explicit mass-conservation, and includes precipitation partitioning, canopy interception, evapotranspiration, snow storage and melt, upper-soil water dynamics, and lower-soil/baseflow processes. Future versions will also incorporate explicit energy-conservation, and the interrelationship between mass and energy, as appropriate.

[10] The GCHM model therefore combines a process-oriented architecture in which water storage and transfer are explicitly represented with the learning capability of ML. This structure provides the basis for identifying the dominant pathways through which precipitation contributes to streamflow and evapotranspiration, and for examining the nature of learned relationships among hydrological processes.

### 1.3.2. But Why Propose Yet Another Catchment Model?

[11] Importantly, our objective is not simply to develop yet another high-performing RER prediction model, but to investigate whether a single proposed (conceptually-interpretable and relatively parsimonious) GCHM architecture can simultaneously provide both (i) accurate streamflow simulations across the diverse variety of hydro-climatic conditions that make up the continental US (CONUS), including semi-arid regions, and (ii) a meaningful basis for hypothesis-driven investigations of internal catchment structure and function across such regions. The term "*Gated*" appended at the beginning of the model name indicates that, although the directed-graph (node-connection) representational architecture of the model is prescribed, the behavioral/functional nature of the process relationships and their dependencies are learnable directly from data.

[12] Further, an important limitation of most available PC-based catchment hydrology models is that they were typically designed to be used with very specific data types/sources having specific model-relevant units. Accordingly, the models can often only be used for locations where (and time-periods when) such model-consistent data sets have been compiled or, conversely the models must be designed specifically to ingest the specific data that are available. For example, to compute evapotranspiration loss from the land surface to the atmosphere, a model may require data that is informative regarding how much capacity the atmosphere has to absorb moisture. In some locations, near-surface air-temperature data may be available from which such information can be derived, whereas in other locations estimates of crop-referenced potential evapotranspiration, or measurements of near-surface relative humidity and wind speed, or even of long and short wave downward radiation may be available, from which information regarding latent and sensible heat fluxes can be derived.

[13] The point is that PC-based models are typically designed to accommodate available data, or the data must be compiled to accommodate available models. As we shall discuss, the "*Gated*" characteristic of the GCHM model enables a single hypothesized model architecture (design of the directed graph of nodes/states and links/fluxes) to be trained and tested with a variety of data sets that may be differently informative (consist of different relevant data types). This ability of the GCHM model to *readily* and *flexibly* ingest different data sources (with retraining), means that it can be used to investigate the informational adequacy of different data sources in a manner that was previously only easily done using purely data-based models. This can help to resolve questions such as whether a model is performing poorly in a particular hydroclimatic regime (such as semi-arid) because of inadequacies in its architecture or due to being provided with poorly informative data.

#### 1.3.3. Benefits of This Approach

[14] Our results, reported here, suggest that simultaneously meeting both of the above-mentioned objectives of accuracy and hypothesis-driven learning is possible. We evaluate predictive performance, internal behavior, and learned flux relationships of the GCHM architecture at 50 catchments representing contrasting hydro-climatic conditions. In particular, we investigate the learned representations of evapotranspiration, snow dynamics, water-storage processes, and flow pathways, and compare the resulting architecture with an LSTM-based data-driven approach. Our findings suggest that *five* state variables are the minimum necessary to achieve a general comprehensive (location agnostic) representation of catchment functioning, with fewer state variables (a pruned model representation) being appropriate at locations where specific hydro-meteorological processes are either absent or can be reasonably ignored while maintaining behavioral consistency.

[15] In summary, this study explores how physically-interpretable computational units can be assembled into a hydro-meteorologically-meaningful catchment-scale modeling framework that can support *both* accurate hydro-meteorological prediction and the hypothesis-driven development of process understanding. It provides a foundation upon which future human-interpretable models that represent the coupled mass-and-energy conserving dynamics of geoscientific systems can be constructed, while enabling the learning of (possibly location-dependent) flux-process relationships from both theory and data.

#### 1.3.4. Contributions and Scope

[16] Specifically, we:

- Formulate a mass-conserving GCHM architecture representing precipitation partitioning, canopy interception, snow dynamics, soil-water storage, evapotranspiration, and baseflow processes.
- Evaluate the model's streamflow prediction performance across catchments with contrasting hydroclimatic conditions.
- Investigate alternative representations of evapotranspiration and their effects on model performance and hydrological behavior.
- Assess the representation of snow storage and melt using independent snow observations.

- Examine the contribution of individual architectural components through systematic ablation experiments and compare the GCH Model with LSTM-based models.
- Interpret the learned states, gates, and flow pathways to investigate whether the model provides physically meaningful insights into catchment rainfall–runoff dynamics.
- Examine how model performance can depend on access to different data types/sources having varying degrees of hydro-meteorological informativeness.

[17] The next section describes the theoretical foundation of the GCH Model, with its explicit focus on catchment *hydrology*.

## 2. The Gated Catchment Hydrology Model (GCHM)

[18] The architecture and system equations of the *Gated Catchment Hydrology (GCH) Model* are based on the Mass-Conserving Perceptron concept introduced by *Wang and Gupta (2024a).*

### 2.1. The Mass Conserving Perceptron (MCP) Unit and its Desirable Properties

[19] The mass-conserving perceptron (MCP) is an ML-based computational unit designed to explicitly enforce physical consistency at each node of an MCP-based neural network. It is structurally analogous to a single unit within a gated recurrent neural network, which allows it to embed the mass-conservation principle directly into data-driven modeling approaches. At the node level, it describes system dynamics through a discrete-time mass balance equation:

$$X_{t+1} = X_t + U_t - O_t - L_t$$

[20] In this formulation, the system state $X_t$ at the next time step at each node is determined by adding incoming mass and subtracting outgoing fluxes over the current interval. When applied to lumped catchment-scale RER modeling, the input term $U_t$ represents the mass flux entering the node (such as incoming precipitation or a mass flux leaving another node), while the output $O_t$ and loss $L_t$ terms correspond to mass fluxes (such as streamflow and evapotranspiration, respectively) leaving the node. The outgoing fluxes are assumed to scale with the current system state according to: $O_t = G_t^O X_t$; $L_t = G_t^L X_t$, where $G_t^O$ and $G_t^L$ are time-dependent gating coefficients that regulate how much mass leaves the node through each pathway. Substituting these relationships into the mass balance equation yields:

$$X_{t+1} = X_t + U_t - G_t^O \cdot X_t - G_t^L \cdot X_t$$

which can be rearranged into a more compact form:

$$X_{t+1} = G_t^R \cdot X_t + U_t$$

where, $G_t^R$ denotes a retention (or "remember") factor that determines the fraction of the state preserved between time steps.

[21] Importantly, these gating coefficients $G_t^O$ and $G_t^L$ are not fixed; instead, they are functions whose values evolve dynamically in response to "*context*", as determined by the system's broader hydro-meteorologic conditions. This context can include information regarding various current and past system states and fluxes computed by the model, and also regarding other environmental variables (such as precipitation, temperature, etc.) provided as data. In terms of notation, we shall together refer to these additional contextually informative variables generically as $\overline{\boldsymbol{C}} = (C_t^1, \dots C_t^M)$ (where each $C_t^i$ refers to a different variable).

[22] This context dependence allows the model to capture complex, time-varying behaviors beyond those that static values for the gating coefficients can represent. Further, they relate to traditional process-based flux equation formulations via the relationships:

$$O_t = G_t^O \cdot X_t = G^O(X_t|\overline{\boldsymbol{C}}^O) \cdot X_t = f_O(X_t|\overline{\boldsymbol{C}}^O)$$
$$L_t = G_t^L \cdot X_t = G^L(X_t|\overline{\boldsymbol{C}}^L) \cdot X_t = f_L(X_t|\overline{\boldsymbol{C}}^L)$$

where $f_O(.)$ and $f_L(.)$ represent the process-based output and loss flux equations whose nature we seek to learn. Therefore, having trained the MCP-based model, the learned data-based representations for $f_O(.)$ and $f_L(.)$, including the nature and strength of their contextual dependence on current and past system states and other environmental variables) can be recovered for examination and/or theory/hypothesis-based modification and replacement. As such, each gating function can be thought of as an embedded, learnable, mini neural network.

[23] Further, from a learning perspective, an important property of the gating functions is that each is constrained to lie within the interval $[0,1]$ to maintain physical validity. Moreover, nodal conservation of mass requires that the gating coefficients at each node must sum to unity: $G_t^R + G_t^O + G_t^L = 1$. This constraint ensures that, at any time step, the entire system state is partitioned into retained, released, or lost components. As a result, the remember gate can be expressed as: $G_t^R = 1 - G_t^O - G_t^L$ which tightly couples the three gating terms.

[24] Starting from an initial condition and a sequence of inputs, the system state can be updated iteratively using this framework, provided the time-varying gating functions are specified. Because these functions are generally unknown, *Wang and Gupta (2024a)* proposed estimating them directly from data using machine learning, enabling flexible yet physically consistent representations of system dynamics.

[25] Moreover, as discussed by (*Wang & Gupta, 2024a, 2024b, 2025*) the MCP unit possesses several key characteristics that make it well suited for interpretable, physically-grounded, conceptual modeling of dynamical systems, such as those encountered in the geosciences (e.g., rainfall–runoff processes). These include:

1) A recurrent structure that captures the Markovian temporal evolution of the system state, allowing memory effects to be explicitly represented.
2) The ability to enforce conservation laws directly at the node level, ensuring physically consistent system behavior.
3) The capacity to account for, and learn, unobserved mass exchanges, including gains and losses that are not directly measured.
4) The flexibility to learn context-dependent process relationships, whereby the gating functions governing system dynamics adapt based on current and antecedent conditions.
5) Straightforward implementation and efficient training using gradient-based optimization (backpropagation), supported by standard machine learning frameworks such as PyTorch.

[26] To this, we add:

6) The ability to learn gating functions (and hence process-flux relationships) that are specific to the kinds of data that are available at any specific location.

[27] In the following section, we describe how the MCP framework can be used to develop a simplified yet process-representative**,** physical-conceptual rainfall–runoff model at the catchment scale, incorporating representations of the key processes of the hydrologic cycle including evapotranspiration (ET), snow water equivalent (SWE), and streamflow.

### 2.2. Rationale for Proposed Model Architecture and System Equations

[28] *Wang and Gupta (2024b)* suggested that up to five state variables may be required to properly represent catchment-scale hydrologic storage behavior, and to capture the multiple timescales underlying rainfall-runoff process dynamics. They also explored various relatively simple MCP-based nodal architectures, both "*physical-conceptual*" and "*network-like*".

[29] For our *Gated Catchment Hydrology Model* (abbreviated as GCHM), we pursue the direction of enhanced scientific interpretability, by developing a *principled* five-state architecture structured around three physically-recognizable land-surface process domains – canopy interception storage, upper soil layer dynamics, and baseflow generation mechanisms – and the partitioning of precipitation into liquid water and frozen snow water equivalent. In other words, we explicitly distinguish between "*surface*" processes (that occur on or above the land surface), "*sub-surface*" processes (that occur below the land surface), and "*interconnecting*" processes

that connect the previous two. We further distinguish the "*surface*" processes into those related to snow on the ground, and those related to frozen (snow/ice) and liquid water interception by vegetation canopy. Finally, we distinguish the "*sub-surface*" processes into those occurring near the land surface and those occurring at greater soil depths. This results in the model architecture shown in **Figure 1**.

[30] In the following sections we explain the different components of the model from precipitation partitioning, to intermediate flow dynamics, to stream flow generation.

#### 2.2.1. Landcover Based Precipitation Partitioning

[31] At the onset of a rainfall event, incoming precipitation $P_t$ is partitioned into two primary components: the fraction intercepted and retained by the vegetation canopy, denoted as $P_t^C$, and the remaining fraction that reaches the ground surface and contributes to soil processes, denoted as $P_t^S$. This partitioning represents a key control on the rainfall-runoff transformation (*Gharari et al., 2021*), as vegetation canopies intercept a portion of precipitation that may subsequently evaporate or drain to the surface. The magnitude of interception is governed by vegetation characteristics – particularly the leaf area index (LAI), which determines the available surface area for water retention.

[32] Following the MCP concept, we model precipitation partitioning using a context-dependent gating function $G_t^{f_{\text{veg}}}$ that controls the fraction of precipitation allocated to canopy storage. The partitioning is expressed as:

$$P_t^C = G_t^{f_{\text{veg}}} \cdot P_t$$

$$P_t^S = (1 - G_t^{f_{\text{veg}}}) \cdot P_t$$

where $P_t$ is the total incoming precipitation at time step $t$.

[33] Further, the vegetation-dependent gating function is defined as:

$$G_t^{f_{\text{veg}}} = 1 - e^{[-a_1 \cdot \text{ReLU}(\widehat{\text{LAI}}(t) - b_1)]}$$

where the context variable $\widehat{\text{LAI}}(t)$ represents scaled leaf area index, obtained using min-max normalization to ensure consistent magnitude across different conditions. The parameters $a_1$ and $b_1$ are learnable coefficients that control the sensitivity and threshold of canopy interception. To ensure physical consistency, both parameters are constrained to be strictly positive ($a_1 > 0, b_1 > 0$). Instead of directly learning the constrained values $a_1$ and $b_1$, we facilitate machine-learning-based training by introduce unconstrained parameters $\hat{a}_1$ and $\hat{b}_1$ that are initialized randomly (from a zero-mean standard normal distribution ($\theta \sim \mathcal{N}(0,1)$)) and recover the constrained values via the transformations $a_1 = \exp(\hat{a}_1)$, $b_1 = \exp(\hat{b}_1)$, thereby enforcing positivity. The ReLU function ensures that interception initiates only when vegetation density exceeds a threshold, while the exponential form provides a smooth, bounded response between 0 and 1.

[34] This formulation reflects the observed physical behavior that denser vegetation leads to greater interception, while sparser vegetation allows most precipitation to reach the soil. Although, in reality, precipitation may be further partitioned into interception, throughfall, and stemflow, the present approach adopts a simplified two-component representation to maintain parsimony while capturing the dominant hydrologic controls.

#### 2.2.2. Temperature Based Precipitation Partitioning

[35] Following canopy interception, the retained precipitation is further partitioned based on air temperature into rainfall and snowfall components. This step accounts for the phase (liquid or frozen) of the precipitation, which significantly influences subsequent hydrological processes such as infiltration, storage, and runoff generation. The partitioning is governed by a temperature-dependent gating function $G_t^{f_{\text{rain}}}$, defined using the sigmoid function as:

$$G_t^{f_{\text{rain}}} = \sigma(a_2 \hat{T}_{\text{air}}(t) - b_2)$$

Where the context variable $\hat{T}_{\text{air}}(t)$ represents normalized air temperature at time step $t$, obtained using min–max scaling. The parameters $a_2$ and $b_2$ are learnable coefficients that control the sensitivity and threshold of the rain–snow partitioning. To ensure physical consistency, $a_2$ is constrained to be positive by introducing an unconstrained parameter $\hat{a}_2$ such that $a_2 = \exp(\hat{a}_2)$, thereby enforcing $a_2 > 0$ during training.

[36] Using this gating function, the canopy-retained precipitation $P_t^C$ is partitioned into rainfall and snowfall components as:

$$P_t^{C\text{Ret: rain}} = G_t^{f\text{rain}} \cdot P_t^C$$
$$P_t^{C\text{Ret: Snow}} = (1 - G_t^{f\text{rain}}) \cdot P_t^C$$

[37] This formulation ensures a smooth and continuous transition between snowfall and rainfall, reflecting the physical behavior that the fraction of each phase of precipitation varies gradually with temperature. At lower temperatures, the gating function approaches zero, leading to precipitation falling predominantly as snowfall, while at higher temperatures it approaches one, resulting in rainfall-dominated conditions.

#### 2.2.3. Canopy Interception Processes

[38] The storage of canopy-intercepted water is represented using two limited-capacity storage units (perceptron tanks) corresponding to snow and rain components, with states $X_t^{C\text{snow}}$ and $X_t^{C\text{rain}}$. These storages are fed by the partitioned canopy-retained precipitation: $X_t^{C\text{snow}} \leftarrow P_t^{C\text{Ret:snow}}$ and $X_t^{C\text{rain}} \leftarrow P_t^{C\text{Ret:rain}}$. A unidirectional snowmelt flux transfers melting snow water from the canopy snow storage to the canopy rain storage:

$$O_t^{C\text{melt}} = G_t^{C_{melt}} \cdot X_t^{C\text{snow}}$$

where the melt gating function is defined as

$$G_t^{C_{melt}} = \sigma(a_3\hat{T}(t) + b_3\widehat{\text{DSW}}(t) - c_3)$$

With the context variables $\hat{T}(t)$ and $\widehat{\text{DSW}}(t)$ denoting normalized air temperature and incoming shortwave radiation, and $a_3$, $b_3$, $c_3$ being learnable parameters controlling the sensitivity and threshold of snowmelt.

[39] Intermediate storage states after precipitation inputs, snowmelt inputs and outputs, and evapotranspiration losses are computed via the operator splitting approach as:

$$X_t'^{C\text{snow}} = X_t^{C\text{snow}} + P_t^{C\text{Ret,Snow}} - O_t^{C\text{melt}}$$
$$X_t'^{C\text{rain}} = X_t^{C\text{rain}} + P_t^{C\text{Ret,Rain}} + O_t^{C\text{melt}} - L_t^{C\text{rain}}$$

where $L_t^{C\text{rain}}$ represents evapotranspiration (ET) loss from canopy computed as $L_t^{C\text{rain}} = G_t^{C_L} \cdot X_t^{C\text{rain}}$ such that:

$$G_t^{C_L} = \sigma(a_4\hat{X}_t^{C\text{rain}} + b_4\widehat{\text{DSW}}_t + c_4\hat{\text{T}}_t + d_4\widehat{\text{VPD}}_t + e_4\hat{u}_t + f_4\hat{\Delta}_t + g_4)$$

Where the context variable $\hat{X}_t^{C\text{rain}}$ represents the canopy rain storage state normalized by its maximum capacity $C_{\max}$, and similarly $\widehat{\text{DSW}}_t$, $\widehat{\text{VPD}}_t$, $\hat{u}_t$, and $\hat{\Delta}_t$ denote the normalized incoming shortwave radiation, vapor pressure deficit, wind speed, and slope of the saturation vapor pressure curve, respectively (for details see *Jawad et al. (2025)*).

[40] In Section 2.3 we investigate various alternative formulations for the canopy evapotranspiration loss gating function $G_t^{C_L}$; the basic variant shown above is a sigmoid-based gating function, in which the coefficients $a_4, b_4, c_4, d_4, e_4, f_4$, and $g_4$ are learnable parameters that control the sensitivity of the gating function to each driver.

[41] Overflow from each storage is then calculated using a Softplus function:

$$OF_t^{C_{snow}} = b_5\, ln\left(1 + e^{\frac{X_t'^{C_{snow}} - a_5(t)}{b_5}}\right) - b_5\, ln\left(1 + e^{\frac{-a_5(t)}{b_5}}\right)$$

$$OF_t^{C_{rain}} = b_5 \, ln\left(1 + e^{\frac{X_t'^{C_{rain}} - a_5(t)}{b_5}}\right) - b_5 \, ln\left(1 + e^{\frac{-a_5(t)}{b_5}}\right)$$

where $a_5(t)$ is the time-varying LAI-dependent capacity computed as $a_5(t) = C_{\max} \cdot \widehat{\mathrm{LAI}}(t)$, with $C_{\max}$ being the maximum canopy storage capacity modulated by the scaled leaf area index $\widehat{\mathrm{LAI}}(t)$, and $b_5$ controlling the smoothness of the overflow response. Both parameters are constrained to be positive.

[42] Finally, the storage states are updated as:

$$X_{t+1}^{C_{\mathrm{snow}}} = X_t'^{C_{\mathrm{snow}}} - OF_t^{C_{\mathrm{snow}}}$$
$$X_{t+1}^{C_{\mathrm{rain}}} = X_t'^{C_{\mathrm{rain}}} - OF_t^{C_{\mathrm{rain}}}$$

[43] For the canopy snow perceptron we control mass balance by constraining the gating functions such that $G_t^{C_{OF:\,rain}} + G_t^{C_{R:\,rain}} = 1$ where $G_t^{C_{OF:\,rain}} = \frac{OF_t^{C_{rain}}}{X_t'^{C_{rain}} + \varepsilon}$ such that $G_t^{C_{R:\,rain}} = 1 - G_t^{C_{OF:\,rain}}$ and $X_{t+1}^{C_{\mathrm{rain}}} = G_t^{C_{OF:\,rain}} \cdot X_t'^{C_{\mathrm{rain}}}$. Finally we have $X_{t+1}^{C_{\mathrm{snow}}} = G_t^{C_{OF:\,snow}} \cdot X_t'^{C_{\mathrm{snow}}}$. This formulation ensures that the canopy storages account for precipitation, snowmelt, canopy evapotranspirative loss, and canopy overflow in a physically-consistent manner.

#### 2.2.4. Upper Soil Water Dynamics

[44] At the soil surface, two conceptual storage components are defined and represented through the soil snow state variable $X_t^{S_{snow}}$ and the soil water state variable $X_t^S$. Soil snow storage is conceptualized as an infinite-capacity reservoir that governs the dynamics of accumulation and release of water as a function of temperature. This representation enables the model to capture phase changes between solid and liquid water, thereby regulating the timing and magnitude of water input to the soil system.

[45] The soil snow storage receives two principal inputs $OF_t^{C_{snow}}$ and $P_t^{S_{snow}}$. The latter represents the fraction of total precipitation contributing to snow accumulation and is defined as:

$$P_t^{S_{snow}} = \left(1 - G_t^{f\,rain}\right) \cdot P_t^S$$

where $G_t^{f\,rain}$ denotes the rainfall fraction function and $P_t^S$ is the total precipitation moving towards the soil surface at time step $t$. This formulation ensures a temperature-dependent partitioning of precipitation into snowfall and rainfall, which is critical for accurately simulating soil water dynamics in regions influenced by seasonal snow processes. Release of water from soil snow storage is governed by a melt-gate mechanism that controls the phase transition from stored snow to liquid water:

$$G_t^{S_{melt}} = \sigma\left(a_6 \hat{T}(t) + b_6 \widehat{DSW}(t) - c_6\right)$$

Where the context variables $\hat{T}(t)$ and $\widehat{DSW}(t)$ denote normalized air temperature and incoming shortwave radiation at time step $t$, respectively. The parameters $a_6$, $b_6$, and $c_6$ are learnable coefficients that regulate the sensitivity and threshold behavior of the snowmelt process at the soil surface.

[46] Accordingly, the outflow from soil snow storage, representing the contribution of meltwater to the soil system, is expressed as:

$$O_t^{S_{melt}} = G_t^{S_{melt}} \cdot X_t^{S_{snow}}$$

[47] This formulation ensures that the release of water is dynamically controlled by both thermal and radiative forcing, allowing for a physically consistent representation of snowmelt-driven inputs to soil water dynamics. Temporal evolution of the soil snow storage state is governed by a mass balance equation that accounts for all incoming and outgoing fluxes, expressed as:

$$X_{t+1}^{S_{snow}} = X_t^{S_{snow}} + OF_t^{C_{snow}} + P_t^{S_{snow}} - O_t^{S_{melt}}$$

[48] Alternatively, the snow storage dynamics can be equivalently expressed using complementary gating functions constrained within the interval $[0, 1]$. In this formulation, the retention and melt processes are

represented through a partitioning mechanism such that: $G_t^{S_{R:snow}} + G_t^{S_{melt}} = 1$ where $G_t^{S_{R:snow}}$ denotes the retention gate controlling the fraction of water remaining in the snow storage, and $G_t^{S_{melt}}$ represents the melt gate governing the fraction released as outflow.

[49] The effective liquid input to the upper soil storage is derived from the combination of multiple hydrological fluxes representing water available at the soil surface. Specifically, three primary contributions are considered: (i) the meltwater flux released from the snow storage, $O_t^{S_{melt}}$; (ii) the rainfall component of precipitation, defined as $P_t^{S1_{rain}} = G_t^{f_{rain}} \cdot P_t^S$; and (iii) the overflow from the canopy rain storage, $OF_t^{C_{rain}}$. These fluxes are aggregated to define total incoming water to the upper soil layer as:

$$P_t^{S3} = O_t^{(S_S)} + P_t^{(S1_{rain})} + OF_t^{(C_{rain})}$$

[50] The first regulatory mechanism acting on this combined input is the infiltration-excess process, which is formulated using a smooth softplus-based gating function. This function partitions the incoming water into two components: (i) an excess fraction that contributes directly to surface runoff, and (ii) a retained fraction that infiltrates into the soil storage $X_t^S$. The excess water is computed as:

$$P_t^{S3_{Excess}} = b_7 \ln\left(1+e^{\frac{P_t^{S3}-a_7}{b_7}}\right) - b_7 \ln\left(1+e^{\frac{-a_7}{b_7}}\right)$$

while the retained (infiltrating) portion is given by:

$$P_t^{S3_{Ret}} = P_t^{S3} - P_t^{S3_{Excess}}$$

[51] Here, $a_7$ and $b_7$ are learnable parameters controlling the precipitation threshold and smoothness of the infiltration-excess response, respectively. This formulation ensures a continuous and differentiable representation of infiltration capacity exceedance, enabling a realistic transition between infiltration-dominated and runoff-dominated conditions.

[52] Next, the upper soil storage is regulated by four principal, gate-controlled fluxes that govern water losses and redistribution. These include: (i) evapotranspiration loss, $L_t^S$, controlled by the gate $G_t^{SL}$; (ii) seepage to the baseflow storage, $O3_t^S$, regulated by $G_t^{SO3}$; (iii) a state-dependent outflow flux, $O2_t^S$, controlled by $G_t^{SO2}$; and (iv) saturation-excess runoff, $O1_t^S$, governed by the gate $G_t^{SO1}$.

[53] The first three fluxes — evapotranspiration ($L_t^S$), seepage to the baseflow storage ($O3_t^S$), and state-dependent outflow ($O2_t^S$) — are formulated as functions of the upper soil moisture state at the current time step, $X_t^S$, thereby reflecting their dependence on available storage. Following their computation, the upper soil moisture state is updated by accounting for both incoming water (i.e., the retained infiltration flux ($P_t^{S3_{Ret}}$) and the outgoing fluxes, yielding an intermediate storage state $X_t'^S$. The saturation-excess runoff flux, $O1_t^S$, is subsequently evaluated based on this intermediate state, ensuring that any excess water generated after internal redistribution is consistently routed as surface runoff.

[54] Given that multiple outgoing fluxes are simultaneously derived from the same storage state, it is essential to enforce mass conservation within the system. To this end, the model constrains the associated gating functions such that the sum of the Output, Loss, Seepage, and Retention (Remember) gates equals unity, i.e.,

$$G_t^{SO2} + G_t^{SL} + G_t^{SO3} + G_t^{SR} = 1$$

This constraint is implemented through a SoftMax-type normalization, specifically:

$$\Psi = \exp(S_{O2}) + \exp(S_L) + \exp(S_{O3}) + \exp(S_R)$$

where $S_{O2}, S_L, S_{O3}$, and $S_R$ are learnable (randomly initialized) parameters. The normalized gate weights are then defined as: $\kappa_{SO2} = \frac{\exp(S_{O2})}{\Psi}, \kappa_{SO3} = \frac{\exp(S_{O3})}{\Psi}, \kappa_{SL} = \frac{\exp(S_L)}{\Psi}, \kappa_{SR} = \frac{\exp(S_R)}{\Psi}$. The individual gating functions are subsequently expressed as:

$$G_t^{SO2} = \kappa_{SO2} \cdot \sigma\left(a_8 \hat{X}_t^S - c_8\right), \text{ such that } O2_t^S = G_t^{SO2} \cdot X_t^S$$

$$G_t^{SO3} = \kappa_{SO3} \cdot \sigma\left(a_9 \hat{X}_t^S - c_9\right), \text{ such that } O3_t^S = G_t^{SO3} \cdot X_t^S$$

$$G_t^{SL} = \kappa_{SL} \cdot \sigma\left(a_{10}\hat{X}_t^S + b_{10}\widehat{DSW}_t + c_{10}\hat{T}_t + d_{10}\widehat{VPD}_t + e_{10}\hat{u}_t + f_{10}\hat{\Delta}_t + g_{10}\right)$$

such that $L_t^S = G_t^{SL} \cdot X_t^S$, where the context variable $\hat{X}_t^S = X_t^S / S_{\max}$ denotes normalized soil moisture state, and the remaining context variables represent normalized meteorological drivers. The intermediate soil moisture state is then computed as:

$$X_t'^S = X_t^S + P_t^{S3_{Ret}} - (O2_t^S + O3_t^S + L_t^S)$$

[55] Based on this intermediate state, the saturation-excess runoff is calculated using a smooth threshold (softplus) formulation:

$$O1_t^S = b_{11}\ln\left(1 + e^{\frac{X_t'^S - a_{11}}{b_{11}}}\right) - b_{11}\ln\left(1 + e^{\frac{-a_{11}}{b_{11}}}\right)$$

where parameter $a_{11}$ is basically $S_{\max}$. Finally, the soil moisture state is updated for the next time step as:

$$X_{t+1}^S = X_t'^S - O1_t^S$$

[56] This formulation ensures a physically consistent and fully differentiable representation of soil water dynamics, while strictly maintaining mass conservation across all flux components.

### 2.2.5. Baseflow Dynamics

[57] From the upper soil storage (with state variable $X_t^S$), water is transferred to the lower soil layer through a unidirectional seepage flux $O3_t^S$. This flux serves as the primary input to the lower soil storage, which is conceptualized as an infinite-capacity reservoir represented by the state variable $X_t^B$.

[58] The lower soil storage is governed by two principal outgoing fluxes: (i) a state-dependent outflow $O_t^B$, which can be interpreted as baseflow and is controlled by the gating function $G_t^{BO}$; and (ii) evapotranspiration loss $L_t^B$, regulated by the gating function $G_t^{BL}$. Consistent with the mass-conserving framework introduced in Section 2.2.4, a SoftMax-based normalization is employed to ensure that all flux partitions from this storage sum to unity. Specifically, we define:

$$\Psi = \exp(B_O) + \exp(B_L) + \exp(B_R)$$

where $B_O$, $B_L$, and $B_R$ are learnable (randomly initialized) parameters. The corresponding normalized weights are given by: $\kappa_{BO} = \frac{\exp(B_O)}{\Psi}, \kappa_{BL} = \frac{\exp(B_L)}{\Psi}, \kappa_{BR} = \frac{\exp(B_R)}{\Psi}$. The gating functions are then formulated as:

$$G_t^{BL} = \kappa_{BL} \cdot \sigma\left(a_{12}\hat{X}_t^B + b_{12}\widehat{DSW}_t + c_{12}\hat{T}_t + d_{12}\widehat{VPD}_t + e_{12}\hat{u}_t + f_{12}\hat{\Delta}_t + g_{12}\right)$$

such that $L_t^B = G_t^{BL} \cdot X_t^B$, and

$$G_t^{BO} = \kappa_{BO} \cdot \sigma\left(a_{13}\hat{X}_t^B - c_{13}\right),$$

such that $O_t^B = G_t^{BO} \cdot X_t^B$.

[59] Since the lower soil storage is treated as an infinite-capacity reservoir, its state is normalized using the maximum capacity of the upper soil storage, i.e., $\hat{X}_t^B = X_t^B / S_{\max}$. The gating functions satisfy the mass-conservation constraint: $G_t^{BO} + G_t^{BL} + G_t^{BR} = 1$, where $G_t^{BR}$ represents the retention (storage) gate. Accordingly, the state update for the lower soil storage is expressed as:

$$X_{t+1}^B = G_t^{BR} \cdot X_t^B = X_t^B - (O_t^B + L_t^B)$$

This formulation ensures a consistent and mass-conserving representation of baseflow generation and evapotranspiration losses from the lower soil layer.

### 2.2.6. Catchment Outflow

[60] Finally, the total outflow flux by the model is governed by the equation:

$$O_t = P_t^{(S3_{Excess})} + O1_t^S + O2_t^S + O_t^B$$

## 2.3. Alternate Formulations of ET

[61] In the formulation presented above, evapotranspiration (ET) losses are represented through three state variables, each corresponding to a distinct hydrological component: canopy storage, soil surface storage, and the lower soil storage. These can be generically denoted individually as $X_t$, with $L_t$ representing the associated canopy interception loss, soil evaporation loss, and ET loss from the lower soil layer. ET loss is formulated as a gated process, where the loss at time $t$ is computed as $L_t = G_t \cdot X_t$, with $G_t$ representing a nonlinear gating function.

[62] In the baseline configuration, this gating function is defined using a sigmoid transformation applied to a combination of meteorological drivers and system states, including air temperature (T), downwelling shortwave radiation (DSW), wind speed (u), vapor pressure deficit (VPD), the slope of the saturation vapor pressure curve (Δ), and the current state $X_t$.

[63] To evaluate alternative formulations, several variants of the gating function were tested (See **Table 1**):

- **$S_1$: Penman-Monteith (PM) with traditional resistances:** In this case, the gating function incorporates ET estimates derived from the Penman–Monteith (PM) equation, where aerodynamic resistance ($r_a$) and surface resistance ($r_s$) are computed using standard empirical formulations (e.g., as described in *Jawad et al. (2025)*). The sigmoid function is applied to a linear combination of $X_t$ and $PM_t^{Trad}$.
- **$S_2$: Penman-Monteith with neural network-based surface resistance:** This formulation retains the Penman–Monteith structure but replaces the empirical surface resistance ($\mathrm{r}_t^s$) with a neural network-based estimate such that the NN takes input the following variables: $DSW_t$, $T_t$, $VPD_t$, $\Delta_t$, and $u_t$ . The resulting ET estimate ($PM_t^{NN}$) is combined with $X_t$ and passed through the sigmoid gating function, allowing for greater flexibility in representing surface controls.
- **$S_3$: Direct meteorological forcing formulation**: The gating function is constructed by directly combining the state variable $X_t$ with meteorological inputs, including net $DSW_t$, $T_t$, $VPD_t$, $\Delta_t$, and $u_t$ followed by a sigmoid transformation. This approach removes reliance on intermediate ET equations and directly links environmental drivers to ET response.
- **$S_4$: Fully data-driven neural network formulation**: In this approach, all input variables (as stated in $S_3$) are passed through a feedforward neural network to produce a learned latent representation ($FNN_t$). This representation is then combined with the state variable $X_t$ and passed through the sigmoid function. This formulation maximizes flexibility by allowing the model to learn complex, nonlinear interactions among inputs without imposing explicit physical structure.

# 3. Experimental Design

## 3.1. Selection of Study Catchments

[64] To assess performance of the proposed GCHM architecture, we applied the behavioral classification approach introduced by *Jiang et al. (2022)* to select 50 representative catchments from the CAMELS dataset. The selected set spans a diverse range of hydro-meteorological conditions and runoff generation processes (see **Table S1** and **Figure 2**). Forty of these catchments fall into four primary streamflow response categories: (i) "*recent-rainfall"* dominated systems in the western United States (RR-W), where discharge mainly reflects rainfall from the same or previous day; (ii) similar "*recent-rainfall"* dominated systems in the eastern United States (RR-E); (iii) "*historical-rainfall"* dominated systems (HR), where streamflow integrates rainfall over periods from several days to weeks; and (iv) "*snowmelt"* dominated systems, where temperature plays the dominant role in regulating discharge (Snow). The remaining ten catchments correspond to "*arid*" regions (Arid; aridity index AI > 1), allowing an assessment of model performance under water-limited conditions.

## 3.2. Input Forcing and Output (Target) Data

[65] To force the models, we used the latest near-surface atmospheric forcing dataset produced by the N*ational Oceanic and Atmospheric Administration* (NOAA), known as the *Analysis of Record for Calibration* (AORC). This dataset provides hourly meteorological variables at a spatial resolution of 800m (*Fall et al., 2023*). The variables—including air temperature, precipitation, incoming shortwave and longwave radiation, surface

pressure, wind speed, and specific humidity—were obtained from AORC version 1.1 (October 1, 2003, to September 30, 2023) and aggregated to daily values to match the temporal resolution required by the models. Leaf Area Index (LAI) was taken from the MODIS MOD15A2H (Level 4), a global dataset with 500 m spatial resolution and an 8-day temporal frequency (*Myneni et al., 2015*). This product has been shown to provide reliable LAI estimates across a wide range of vegetation types (*Yan et al. (2016)*). Surface emissivity data, available at 1 km resolution and 8-day intervals, were obtained from the MODIS MOD21A2 (*Hulley, 2017*). In addition, hourly albedo data at 9 km resolution were sourced from the ERA5-Land reanalysis (*Muñoz-Sabater et al., 2021*), developed by the European Centre for Medium-Range Weather Forecasts. All variables were spatially averaged over each catchment prior to analysis.

[66] Observed daily streamflow, used as the target variable, was obtained for all 15 catchment outlets from the *United States Geological Survey* (USGS) via its online data portal (*USGS, 2025*). These records are reported in cubic feet per second ($ft^3/s$) as mean daily discharge at each gauging station. Discharge was converted to millimeters per day by transforming flow into daily volume and normalizing by the corresponding catchment area, resulting in area-standardized streamflow suitable for analysis.

### 3.3. Metrics Used for Training and Performance Assessment

[67] The metric used to train all the models used in this study is the *Jawad-Kling-Gupta Efficiency* ($JKGE_{ss}$; *Jawad et al. 2026*) that improves information extraction from the data (and hence model performance) by explicitly accounting for non-stationarity in the streamflow conditional-data-generating-process. *Jawad et al. (2026)* showed that use of $JKGE_{ss}$ for training of both physical-conceptual and data-based dynamical systems models results in more robust evaluation performance across diverse hydro-meteorological regimes and levels of model adequacy than metrics such as the *Nash–Sutcliffe Efficiency* (*Nash & Sutcliffe, 1970*) and classical *Kling–Gupta Efficiency* (*Gupta et al., 2009; Gupta & Kling, 2011*), or in particular the skill-score adjusted version of the latter ($KGE_{ss}$; (Knoben et al., 2019)), all of which do not recognize and account for such non-stationarity.

[68] The $JKGE_{ss}$ metric is constructed from four statistically- and physically-interpretable components $M$, $M^*$, $V^*$, $C^*$ (Eqns 1-5) as follows:

$$JKGE_{ss} = 1 - \sqrt{(M + M^* + V^* + C^*)/2} \quad \text{(Eq. 1)}$$

$$M = (1 - \beta_{so})^2 \quad \text{(Eq. 2)}$$

$$M^* = \frac{1}{N}\sum_{t=1}^{N}\left(1 - \frac{b_t^s}{b_t^o}\right)^2 \quad \text{(Eq. 3)}$$

$$V^* = (1 - \alpha_{so}^*)^2 \quad \text{(Eq. 4)}$$

$$C^* = (1 - \rho_{so}^*)^2 \quad \text{(Eq. 5)}$$

where

- $M$ is the time-stationary long-term mass balance component (optimal value zero): $\beta_{so} = \frac{\mu_s}{\mu_o}$ such that $\mu_s$ and $\mu_o$ are long-term means of the simulated and observed system output timeseries (here being streamflow).
- $M^*$ is the time non-stationary anomaly mass balance component (optimal value zero): $b_t^o = f_b(t|\bar{O})$ and $b_t^s = f_b(t|\bar{S})$ represent time-varying benchmark time series values obtained by applying a pre-specified functional operation $f_b(.)$ to the observed signal $\bar{O}$ and the simulated signal $\bar{S}$, respectively.
- $V^*$ is the anomaly variance component (optimal value zero): $\alpha_{so}^* = \frac{\psi_s}{\psi_o}$ where $\psi_s = \sqrt{\frac{1}{N}\sum_{t=1}^{N}(s_t - b_t^s)^2}$ and $\psi_o = \sqrt{\frac{1}{N}\sum_{t=1}^{N}(o_t - b_t^o)^2}$, represents the standard deviations of the benchmark-centered observed and simulated "*anomaly*" time-series (obtained by subtracting the corresponding time-non-stationary benchmarks).

- $C^*$ is the anomaly cross correlation component (optimal value zero): $\rho_{so}^* = \frac{1}{N}\sum_{t=1}^{N}\left(\frac{s_t - b_t^s}{\psi_s}\right)\cdot\left(\frac{o_t - b_t^o}{\psi_o}\right)$.

[69] The optimal $JKGE_{ss} = 1$ efficiency value is obtained when the component terms $M$, $M^*$, $V^*$ and $C^*$ are all zero, while the benchmark $JKGE_{ss} = 0$ corresponds to the model performance being no better than reproducing *both* the long-term water balance (i.e., $M = 0$) and the time-varying seasonal and interannual trends in water balance (i.e., $M^* = 0$), due to which we would have $V^* = C^* = 1$. Accordingly, $JKGE_{ss}$ is a *stronger* metric of model performance than $KGE_{ss}$, which only requires reproducing both the long-term water balance (i.e., $M = 0$) to achieve its benchmark value of $KGE_{ss} = 0$. Negative values of $JKGE_{ss}$ indicate poor reproduction of long-term and time-varying water balance.

[70] Following the detailed analysis of CONUS wide catchments from different hydro-climatic regimes conducted by *Jawad et al. (2026)*, we compute the observed benchmark timeseries ($b_t^o$) as the "*sectional mean*" of length $N_s = 30$ days such that $b_t^o = \frac{1}{N_s}\sum_{i\,\in\,G(t)} O_i$ where $G(t)$ indicates the non-overlapping section to which the particular time step $t$ belongs, and $O_i$ indicates the observed timeseries within that section. Similarly, the simulated benchmark timeseries is computed as $b_t^s = \frac{1}{N_s}\sum_{i\,\in\,G(t)} S_i$ with $N_s$=30 days.

## 3.4. Benchmark Models

### 3.4.1. Ablated Versions of the GCH Model

[71] We refer to the five-state GCHM architecture using the notation $GCHM_5$. In this study we also evaluate the ablated versions of $GCHM_5$ listed below, to assess the importance of individual physical processes in contributing to overall model performance (**Figure 3**). To do this, key components of the full physical-conceptual structure were selectively removed, and the effect on model performance was analyzed across catchments and hydro-climatic regimes. This approach allows us to quantify the relative importance of different process representations and to identify which mechanisms are most critical for maintaining robust and consistent model behavior under varying hydro-climatic conditions.

a) $\boldsymbol{GCHM_2}$: This two-state architecture ignores canopy-related processes, and can be interpreted as representing the temporal dynamics of snow water equivalent ($X_t^{S_{snow}}$) and soil storage ($X_t^S$) in a highly parsimonious manner. The model incorporates three parallel flow pathways that collectively contribute to the total system outflow ($O_t$): one for infiltration excess ($P_t^{S2_{Excess}}$), another for soil saturation excess ($O1_t^S$), and the third one for state dependent outflow flux ($O2_t^S$). Conceptually, $GCHM_2$ represents the most basic hydrological processes of snow accumulation/melt and soil water dynamics, while allowing runoff generation through both infiltration-driven and saturation-driven mechanisms.

b) $\boldsymbol{GCHM_3^{NC}}$: This three-state architecture augments $GCHM_2$ with a baseflow storage ($X_t^B$) and a seepage flux that connects the soil state ($X_t^S$) to the baseflow state ($X_t^B$) through a one-directional flow path; the superscript NC refers to "*no canopy*". The additional pathway enables clearer separation between quicker surface and near-surface flow runoff processes and slower subsurface drainage. The inclusion of baseflow storage allows $GCHM_3^{NC}$ to better represent delayed flow contributions and to improve hydrograph recession behavior, while still maintaining a relatively parsimonious representation.

c) $\boldsymbol{GCHM_3^{NS}}$: This alternative three-state architecture differs from $GCHM_3^{NC}$ by removing the snow state variable ($X_t^S$) and reintroducing the canopy water storage state ($X_t^C$); the superscript NS refers to "*no snow*". This modification alters the internal process representation by excluding snow accumulation and melt dynamics, while introducing canopy interception and storage processes. The architecture is implemented to systematically evaluate how the absence of an explicit snow component affects model behavior and predictive performance, particularly in terms of its ability to capture seasonal variability, runoff generation, and overall hydrologic response across different catchment conditions.

### 3.4.2. Pure ML-Based Upper Benchmark Models

[72] To assess and compare the performance of the $GCHM_5$ model against pure ML-based models, we include catchment-specific long short-term memory (LSTM-based) network models of varying complexity. Important in

the context of our study is the fact that LSTM-based models are also "gated" models (belonging to the class of recurrent neural networks) and have been shown to be highly effective for data-driven modeling of dynamical systems (*Kratzert et al., 2018*). Their strength lies in their cell-state structure and their ability to learn context-dependent gating mechanisms that can exploit short- and long-term dependencies (*Hochreiter & Schmidhuber, 1997*). Following the work by *(Wang & Gupta, 2024a, 2024b)*, single-layer LSTM architectures with up to 5 hidden nodes were tested in this study.

### 3.5. Data Splitting Method used for Model Development

[73] Robust model development requires dividing the available dataset $D$ into separate subsets for calibration and evaluation, ensuring that model evaluation is conducted on independent data. Following the approach introduced by *(Zheng et al., 2022)*, the 20-year record was partitioned into training ($D_{\text{Train}}$) and evaluation ($D_{\text{Eval}}$) datasets using a year-based stratified sampling strategy. In this approach, all years were first ranked according to their total annual streamflow. The pair of most extreme years—representing the wettest and driest conditions—were initially assigned to the training set, while the next pair of extreme years was allocated to the evaluation set. After removing these selected years, the procedure was repeated iteratively until all years were assigned. Ultimately, 60% of the year pairs (12 water years) were included in $D_{\text{Train}}$, with the remaining 40% assigned to $D_{\text{Eval}}$. This stratification method ensures that both subsets encompass a comparable spread of hydroclimatic variability, thereby reducing the risk of bias and improving the representativeness of model training and evaluation.

### 3.6. Hyper-Parameter and Parameter Training Procedure

[74] The training procedure for all of the models generally followed the framework outlined by *Wang and Gupta (2024a)*. To mitigate sensitivity to uncertain initial conditions in the PC-based models, a three-year spin-up period was introduced by repeating the first water year (WY 2004) three times at the start of the simulation. As discussed by *De la Fuente et al. (2023)*, this approach reduces the impact of incorrect state initialization. All models were implemented in PyTorch, with parameter optimization carried out using the Adam optimizer (*Kingma, 2014*). The loss function and corresponding gradients were computed over the training subset. For each catchment, and for every model architecture, models were trained using 10 different random seeds to account for variability in parameter initialization. Each training run was conducted for 1500 epochs using full-batch updates. A learning rate of $1.0 \times 10^{-1}$ was adopted, and all trainable parameters were initialized from a zero-mean standard normal distribution ($\theta \sim \mathcal{N}(0,1)$) to encourage stable optimization in the early stages of training. The final model configuration was selected based on the seed that achieved the best average loss across both $D_{\text{Train}}$ and $D_{\text{Eval}}$. Each model architecture was trained separately for all 50 catchments, allowing for catchment-specific characterization of input–state–output relationships. This design enabled a systematic evaluation of model performance and sensitivity across architectures of varying complexity. Future work will investigate regional model training where model parameters at each catchment are predicted via regionalization relationships, so as to better exploit the power of large sample hydrology (*Gupta et al., 2014; Kratzert et al., 2019*), and to facilitate prediction in ungaged basins (*Kratzert et al., 2018; Kratzert et al., 2019*).

## 4. Experimental Results

### 4.1. Performance Assessment of $GCHM_5$

#### 4.1.1. Performance Assessment via Evaluation Metrics

[75] For the five-state $GCHM_5$ model, **Figure 4** presents box-and-whisker plots of the performance metrics computed over the complete record, with one value per catchment, across the five hydroclimatic regimes. Panel (a) shows the non-stationary metric $JKGE_{ss}$ and its components ($M$, $M^*$, $V^*$ and $\rho^*$) and panel (c) shows the classical stationary metrics $KGE_{ss}$ and its components ($\beta$, $\alpha$, and $\rho$). Panels (b) and (d) map the catchment-level $JKGE_{ss}$ and $KGE_{ss}$ values.

[76] Note that, for $JKGE_{ss}$ and its corresponding correlation term ($\rho^*$), and for $KGE_{ss}$ and its corresponding correlation, mass-balance and variability terms ($\rho$, $\beta$ and $\alpha$), a value of 1.0 represents perfect performance. However, for the $JKGE_{ss}$-related mass-balance and variability error terms ($M$, $M^*$, *and* $V^*$), better performance

is indicated by values closer to zero. Within each panel, each color-coded box plot corresponds to one of the five hydroclimatic regimes defined in **Figure 2**, and its spread reflects the variability in model performance among the catchments of that regime.

[77] Note further that $JKGE_{ss}$ and $KGE_{ss}$ represent the non-stationary and stationary skill-score versions of $KGE$, with $KGE_{ss}$ representing the classical measure commonly reported in the literature (where the benchmark value of zero represents performance no better than simply using the "*long-term mean*" as the prediction). As explained in Section 3.3, $JKGE_{ss}$ is a much <u>*stronger*</u> metric of model performance because the benchmark value of zero represents performance comparable to using the "*time-varying monthly mean*" as the prediction (in other words the model is able to reproduce <u>*both*</u> the long-term mean and the seasonal and interannual trends in monthly streamflow). Accordingly, we can expect $JKGE_{ss} \leq KGE_{ss}$ with equality only in the unlikely case that there is no interannual and seasonal variability to monthly streamflow.

[78] Meanwhile, the $M$ component represents how well the model reproduces the (stationary) long-term mass balance, $M^*$ represents how well the model reproduces the (non-stationary) anomaly mass balance, $V^*$ represents how well the model reproduces the anomaly variability, and $\rho^*$ represents how well the model reproduces the anomaly cross-correlation.

[1] Clearly, the five-state $GCHM_5$ model performs well across all regimes, with median $JKGE_{ss}$ values of roughly 0.7 to 0.85 and median $KGE_{ss}$ values of roughly 0.87 to 0.95. Performance is best and most consistent among *snowmelt* dominated (black) catchments. *Historical-rainfall* (green) and *recent-rainfall (E)* (blue) catchments show the lowest medians and the widest catchment-to-catchment spread while, surprisingly, the model performs remarkably well in the *arid* regime (orange), apart from a few outlier catchments. Model performance in terms of each aspect can be summarized as follows:

- $\boldsymbol{JKGE_{ss}}$: The metric (on which the model was trained) exhibits notable variation across the five hydroclimatic regimes. *Snowmelt* dominated catchments show the highest median (~0.84) and the smallest spread, suggesting that they are the easiest to model. The *arid regime* has the second-highest median (~0.78) and a relatively compact interquartile range, but also several low outliers ($JKGE_{ss} \approx$ 0.5), reflecting a small number of poorly simulated catchments. *Recent-rainfall* (W) has an intermediate median (~0.75), while *historical-rainfall* and *recent-rainfall* (E) have the lowest medians (~0.71–0.72) and the largest spread, with whiskers extending down to ~0.5 and an outlier near 0.4 for *recent-rainfall* (E),. The best-performing catchments reach comparable $JKGE_{ss}$ values across all five regimes (~0.85–0.88), indicating that every regime contains catchments that are modeled well. The map in panel (b) shows the lowest values (~0.3–0.5) at a few arid catchments in the southwest, at historical-rainfall catchments in the south-central US, and at one *recent-rainfall* (E) catchment in Florida.
- $\boldsymbol{KGE_{ss}}$: A similar pattern is observed for this metric (on which the model was <u>*not*</u> trained), but with values that are consistently higher than that of $JKGE_{ss}$, as expected given its relative weakness. *Snowmelt* dominated catchments again perform best (median ~0.95) with a very compact distribution. *Historical-rainfall* and *arid* catchments have medians of ~0.90, followed by *recent-rainfall (W)* (~0.88) and *recent-rainfall (E)* (~0.87). *Recent-rainfall (E)* shows the greatest spread, and the arid regime has one pronounced outlier (~0.5). The large gap between $KGE_{ss}$ and $JKGE_{ss}$ in all regimes shows how much of the apparent skill under the classical metric comes from reproducing the seasonal cycle, and confirms that $JKGE_{ss}$ is a much stronger assessment of model performance.
- $\boldsymbol{M}$ ***and*** $\boldsymbol{\beta}$: The long-term mass-balance error is close to zero for all regimes, with very narrow interquartile ranges, indicating that the model effectively reproduces the long-term water balance in nearly all catchments. Only *historical-rainfall* and *recent-rainfall (E)* show slightly larger errors, with a few outlier catchments (up to ~0.15). Consistent with this, the $KGE_{ss}$ bias ratio β is close to 1 for all regimes (medians ~0.96–0.98), the exception being *historical-rainfall*, where β is slightly lower (~0.92), indicating a mild tendency to underestimate mean flow.
- $\boldsymbol{M^*}$: The anomaly mass-balance error differs more clearly across regimes, indicating its strength as a metric for model discrimination. It is lowest for *snowmelt* dominated catchments (median ~0.07), moderate for *recent-rainfall (W)* and *arid* (~0.12), and highest and most variable for historical-rainfall and *recent-rainfall (E)* (median ~0.22). This indicates that the model has more difficulty reproducing deviations from the seasonal mass balance in the *historical-rainfall* and *recent-rainfall (E)* regimes, consistent with their lower $JKGE_{ss}$ values.

- $\boldsymbol{\alpha}, \mathbf{V}^*, \boldsymbol{\rho}$, ***and*** $\boldsymbol{\rho}^*$: The anomaly variability error $V^*$ is near zero for all regimes, with a slightly higher median and spread for recent-rainfall (E), including one outlier (~0.2). This indicates that the model reproduces the magnitude of flow variations about the non-stationary benchmark well, consistent with α being close to 1 in most regimes (the α spread is largest for the arid regime and also notable for *recent-rainfall* (E)). The anomaly cross-correlation $\rho^*$ therefore appears to also be a major factor behind differences in $JKGE_{ss}$. *Snowmelt* dominated catchments achieve the highest median $\rho^*$ (~0.88), reflecting strong temporal agreement between observed and simulated flow anomalies, which makes sense given the more smoothly-varying snowmelt-driven temporal dynamics of streamflow in such regions. *Historical-rainfall, arid,* and *recent-rainfall (E)* have intermediate medians (~0.82–0.85), and *recent-rainfall (W)* has the lowest median (~0.72) and the widest spread. The same ordering appears in the $\rho$, where snowmelt is highest (~0.95) and *recent-rainfall (W)* and *recent-rainfall (E)* are lowest (~0.85–0.86).

#### 4.1.2. Performance in Reproduction of Streamflow Hydrographs

[79] **Figure 5** provides a more intuitive and qualitative assessment of model performance by visual examination of the streamflow hydrographs. Hydrograph comparisons are shown for five representative catchments (each corresponding to a distinct hydroclimatic regime) for both a wet (panel a) and a dry (panel b) year. Overall, we see that the model demonstrates a strong ability to capture the timing and magnitude of streamflow dynamics across most regimes, with notable differences in performance depending on hydroclimatic conditions.

[80] In wet years:

- The model accurately reproduces peak flows and seasonal patterns in *snowmelt* dominated and *historical-rainfall* catchments, with simulated flows closely matching observations in both magnitude and timing.
- The *recent-rainfall* regimes, particularly in the East, also show reasonable agreement, although some deviations are observed during high-flow events.
- In contrast, the *recent-rainfall (W)* and *arid* catchments exhibit some discrepancies in peak flow representation, indicating challenges in reproducing rapid runoff responses.

[81] Model performance during dry years remains comparable to that observed in wet years across most hydroclimatic regimes, indicating robustness of the model in reproducing streamflow dynamics under varying conditions.

- The *snowmelt* dominated and *historical-rainfall* catchments continue to show good agreement between observed and simulated flows in terms of both timing and magnitude.
- Similarly, both of the *recent-rainfall* regimes maintain a reasonable level of accuracy, although minor deviations persist during certain flow events.
- However, in *arid* catchments, peak flows are overestimated, and discrepancies in representing low-flow conditions are more evident.
- Despite these limitations, overall performance during dry years remains consistent with that observed during wet years, highlighting the model's ability to generalize across contrasting hydrological conditions.

#### 4.1.3. Learned Functional Forms of the Gating Functions

[82] The $GCHM_5$ model architecture contains various gating functions that regulate the fluxes from different state variables. **Figure 6** shows the learned functional forms of these gates. In Panels c-g, the thin lines indicate gating functions for individual catchments, while thick lines show the averages across each hydro-climatic category.

- $\boldsymbol{G^{S_{melt}}}$ **(Panel a)**: This gate controls snowmelt activation as a function of temperature (Temp) and incoming shortwave radiation (DSWR). Consistent with physical understanding, all three of the *snowmelt* dominated catchments shown here have learned a sharp monotonic dependence on temperature, showing steep threshold-like activation in the region of 0-5 °C. In contrast the dependence on DSWR is relatively weak.
- $\boldsymbol{G^{S_{Excess}}}$ **(Panel b):** This softplus-type infiltration excess gate operates when the effective precipitation $P^{S3}$ exceeds a threshold limit ($P_{max}$). It regulates how excess water contributes to runoff/storage. Wet *historical-rainfall* catchments (green) activate early and rise smoothly, meaning that even small amounts of excess generate responses. *Arid* catchments (orange) show delayed and more gradual increases, indicating that larger excess is required for activation. *Snowmelt* dominated basins (black) have the slowest rise, consistent with buffering/storage effects.
- $\boldsymbol{G^{SO1}}$ **(Panel c):** This softplus-type saturation excess gate operates when the upper soil zone state variable $X^S$ exceeds a threshold limit ($S_{max}$). *Arid* catchments (orange) show smooth activation at relatively smaller values of upper soil zone storage. *Recent-rainfall* catchments, both East (blue) and West (red), activate at mid-range values of upper soil zone storage, indicating stronger nonlinear responses once the soils wet up. *Snowmelt* dominated systems (black) activate more gradually and at higher values of upper soil zone storage, reflecting higher storage capacity and slower release.
- $\boldsymbol{G^{SO2}}$ **(Panel d):** This gate represents state dependent outflow flux from the upper soil zone storage, which increases monotonically as the state $X^S$ increases. *Recent-rainfall-West* (red) catchments show sharp threshold-like increase to a constant conductivity rate (slightly above 0.1), indicating activation beyond a certain storage level (~50-150 mm). *Snowmelt* dominated *basins* show more gradual increase in conductivity, while *arid* and *historical-rainfall* systems remain low, indicating very low contributions to streamflow from the upper soil zone.
- $\boldsymbol{G^{SO3}}$ **(Panel e):** This gate represents seepage (recharge) from the upper soil zone storage to the lower soil zone storage, driven by build-up of upper soil zone moisture ($X^S$). The gate value increases from near zero conductivity at low values for upper soil zone storage to higher conductivity as the soil wets up. *Snowmelt* dominated (black) catchments and *recent-rainfall -West* systems (red) activate early (at lower storage thresholds) and more steeply, indicating stronger vertical connectivity. *Arid* catchments (orange) show relatively limited and more gradual activation, suggesting weaker vertical connectivity and/or more restricted seepage pathways.
- $\boldsymbol{G^{BO}}$ **(Panel f):** This gate controls state dependent outflow flux (baseflow due to groundwater outflow) from the lower soil zone as a function of storage ($X^B$). *Snowmelt* dominated basins (black) show a steady, near-linear increase, indicating sustained groundwater contribution. *Arid* catchments (orange) exhibit delayed nonlinear rise, while *recent-rainfall -West* systems (red) remain lower and increase more gradually, indicating weaker baseflow response. *Historical-rainfall* (green) catchments show smooth and moderate rates of growth. Note the order-of-magnitude differences in conductivity values between the lower soil zone (Panel f) and the upper soil zone (Panel d), consistent with hydrologic theory and observation.

#### 4.1.4. Dominant Flow Paths

[83] The $GCHM_5$ model (**Figure 3**) represents four primary flow pathways/components, representing different runoff generation mechanisms, through which water contributes to total streamflow ($O_t$).

- The first component, $P_t^{S3_{\text{Excess}}}$, corresponds to infiltration-excess runoff, generated when rainfall intensity exceeds the infiltration capacity of the upper soil zone.

- The second component, $O1_t^S$, represents saturation-excess runoff, which occurs when the upper soil zone becomes fully saturated and any additional input contributes directly to streamflow.
- The third term, $O2_t^S$, is a state-dependent outflow flux from the upper soil zone, representing lateral near-surface flow processes that depend on the current moisture storage and horizontal state-dependent (unconfined) hydraulic conductivity of the upper soil zone.
- Finally, $O_t^B$ denotes baseflow, which originates from deeper (lower soil zone) groundwater storage and contributes more gradually to streamflow.

[84] While all four pathways contribute to total streamflow, their relative importance and magnitudes of contribution vary across hydro-climatic regime. In some systems, particularly in wetter or more responsive landscapes, fast-streamflow-response mechanisms such as infiltration- or saturation-excess runoff dominate. In contrast, the streamflow response of catchments with larger soil storage capacity or permeability may be dominated by slower pathways such as state-dependent subsurface flow or baseflow. This variability reflects differences in hydroclimatic conditions, soil properties, and catchment structure.

[85] **Figure 7** visualizes, for five representative catchments, the four different flow pathway contributions to streamflow under (a) wet years and (b) dry years, along with (c) the frequency of activation of each pathway across all 50 catchments. In panels (a) and (b), the stacked hydrographs show the temporal contributions of the four flow components to total streamflow. The results indicate that:

- In the *recent-rainfall* (first and fourth rows) and *historical-rainfall* (third row) catchments, the sharp peaks that dominate high-flow events, particularly in wet years (column a), result from a combination of infiltration-excess runoff ($P_t^{S3_{\text{Excess}}}$, magenta) and saturation-excess runoff ($O1_t^S$, cyan), indicating rapid runoff generation driven by intense rates of precipitation. Of course, in the drier years (column b), the reduced contributions from this pathway result in lower peak magnitudes. In the *snowmelt* dominated (second row) and *arid* (fifth row) catchment types the infiltration-excess runoff pathway was not adopted.
- Overall, except in the *snowmelt* dominated catchments, saturation-excess runoff ($O1_t^S$, cyan) is a significant component of streamflow during moderate flow conditions, especially in wetter catchments where saturation of the upper soil zone occurs more frequently. This component shows a more sustained contribution than infiltration-excess runoff, particularly during prolonged wet periods.
- The upper soil zone storage-and-lateral-hydraulic-conductivity-dependent contribution to streamflow ($O2_t^S$, green) is especially evident in the *snowmelt* dominated systems, where stored upper soil zone water is released gradually over time. This component is also significant in the *arid* systems, where it exhibits a delayed and smoother response.
- The baseflow component ($O_t^B$, brown) remains relatively small but persistent, providing a continuous background contribution to streamflow. Its significance is more apparent during low-flow periods and in dry years, where it sustains flow in the absence of significant precipitation input.

[86] Panel (c) of **Figure 7** summarizes the frequency (percent of timesteps) of activation of each pathway, defined as occurrences exceeding 0.01 mm over 7300 timesteps (20 years). The results show that:

- Baseflow and soil outflow pathways ($O_t^B$ and $O2_t^S$) are active for a large fraction of time, particularly in *wetter* and *snowmelt* dominated catchments, indicating their role as persistent background processes.
- In contrast, infiltration-excess runoff ($P_t^{S3_{\text{Excess}}}$) is activated less frequently but contributes disproportionately during peak events, especially in *recent-rainfall* systems.
- *Saturation-excess* runoff ($O1_t^S$) shows intermediate activation frequencies, with higher occurrence in *wetter* regimes and lower in *arid* catchments.

[87] Overall, these results correspond well with conceptual hydrological understanding of how catchments store and partition incoming precipitation in different hydro-climatic regimes, supporting the usefulness of the GCHM model for scientifically interpretable investigation and counterfactual analysis.

#### 4.1.5. Comparison of Model Generated Snow Water Equivalent (SWE) with UA-SWE:

[88] We compared the simulated values of snow water equivalent (SWE) with an independent observational dataset to evaluate the ability of the $GCHM_5$ model to represent snow accumulation and melt dynamics. Model-simulated SWE is computed as the sum of snow water stored in the canopy and on the ground: $\text{SWE}_t = X_t^{C_{\text{snow}}} + X_t^{S_{\text{snow}}}$ where $X_t^{C_{\text{snow}}}$ represents snow stored in the canopy, and $X_t^{S_{\text{snow}}}$ represents snow stored at the ground/soil surface.

[89] The analysis includes five catchments where snowfall contributes either as a primary or secondary source of precipitation – three are classified as *snowmelt* dominated (top three rows), where seasonal snow accumulation and melt strongly control hydrological response, while two are *arid* catchments that receive intermittent snowfall and exhibit weaker influence of snow. For evaluation, we used the gridded University of Arizona Snow Water Equivalent product (UA-SWE; *Broxton et al. (2019)*), derived using remote sensing and data assimilation techniques, that provides spatio-temporally continuous estimates of snow water storage and is widely used for large-scale hydrological and climate studies due to its consistency and coverage.

[90] **Figure 8** compares seven water years (WY-2004 to WY-2010) of model-simulated SWE (blue) and UA-SWE (red) for the five aforementioned catchments. The results show that:

- Overall, despite not being trained to reproduce SWE observations, the model demonstrates strong skill in reproducing SWE dynamics. It captures the seasonal patterns of snow accumulation and melt, and the general timing and magnitude of SWE across all catchments.
- Some discrepancies in peak values and melt rates are observed for the *arid* catchments, likely due to uncertainties in precipitation phase partitioning, temperature thresholds controlling melt, and inability of the spatially-lumped model to represent the effects of spatial heterogeneity.
- Nonetheless, the results highlight the ability of the model to realistically represent SWE dynamics across a wide range of hydroclimatic conditions, even without direct calibration to SWE observations.

#### 4.1.6. Impact of Various ET Formulation on Model Performance

[91] We further evaluated (**Figure 9**) the impact on model performance of using different methods for computing evapotranspiration flux (ET), listed as $S_1$ to $S_4$ (see Section 2.3). Model performance is assessed using boxplots of the $JKGE_{ss}$ metric.

[92] Recall that methods $S_1$ and $S_2$ incorporate ET estimates derived from the Penman–Monteith (PM) equation (where $S_1$ uses standard empirical formulations for aerodynamic resistance and surface resistance, while method $S_2$ uses neural network-based estimates of the surface resistances). Method $S_3$ uses a sigmoidal gating function formulation that uses a linear combination of the corresponding state variable $X_t$ and net $DSW_t$, $T_t$, $VPD_t$, $\Delta_t$, and $u_t$ meteorological data as contextual inputs. Finally, method $S_4$ employs a fully data-driven neural network formulation that maximizes flexibility by allowing the model to learn complex, nonlinear interactions among the inputs without imposing explicit physical structure.

[93] The results indicate that:

- In the *recent-rainfall* systems (red RR-W and blue RR-E), all four formulations perform similarly well. The median values of $JKGE_{ss}$ are consistently high, with only minor variations between formulations. This indicates that although use of $S_1$ results in comparatively worse performance, the improvements from $S_1$ to $S_4$ are marginal, and all formulations are comparably good at contributing to the reproduction of streamflow dynamics.
- In *snowmelt* dominated catchments (black), the median performance gradually improves from $S_1$ to $S_3$, and then shows clear improvement with $S_4$ with significantly better median and minimum values, indicating that model robustness improves with increasing flexibility of the ET formulation.

- In *historical-rainfall* catchments (green), the standard Penman–Monteith (PM)-based formulation $S_1$ performs noticeably worse than the alternative formulations. $S_2$ yields a significant improvement, suggesting that its representation of the resistance terms in the PM equation is more effective. $S_3$ further improves performance relative to $S_1$. However, $S_4$ achieves the best overall performance, significantly outperforming the alternatives.
- The largest differences are observed in *arid* catchments, where performance improves substantially from $S_1$ to $S_4$. For the standard Penman–Monteith (PM)-based $S_1$ formulation, $JKGE_{ss}$ values are generally low with large variability, indicating poor and inconsistent performance. As we move toward $S_4$, there is a clear increase in median performance along with a reduction in spread. This highlights the strong sensitivity of *arid* systems to the manner in which ET is estimated.

[94] We also observe that the traditional PM-based formulation ($S_1$) performs consistently worst across all hydroclimatic regimes, with particularly poor performance in *arid* and *historical-rainfall* systems. In contrast, the more flexible formulations show clear improvements. Further, replacing the empirical formulation of surface resistance ($r_s$) with a neural network-based function in $S_2$ leads to a noticeable improvement over $S_1$; this suggests that empirical resistance parameterizations limit model performance, and introducing flexibility in representing surface controls on evapotranspiration enhances overall model skill. Finally, the $S_4$ formulation performs better across all catchment types, indicating that direct incorporation of meteorological drivers and fully data-driven approaches yield comparable gains in performance.

[95] Overall, these results highlight that increasing flexibility in ET representation, whether through modified physical formulations or data-driven approaches, results in consistent improvements in model performance across hydroclimatic regimes.

## 4.2. Comparison of $GCHM_5$ with Ablated Versions and Pure Data Based Models

[96] Next, we compared the $GCHM_5$ model with its ablated versions (as explained in section 3.4.1), and with purely ML-based models, to investigate how the different conceptual components represented within the newly proposed model architecture contribute to its performance.

### 4.2.1. Comparison using Metric Box Plots

[97] **Figure 10** compares the performance of the full $GCHM_5$ model with its ablated versions ($GCHM_3^{NC}$, $GCHM_3^{NS}$ and $GCHM_2$) and the benchmark purely data-driven five-cell-state LSTM-based model across different hydroclimatic regimes.

- Across all catchments, the five-state $GCHM_5$ architecture consistently shows strong and stable performance, with high median $JKGE_{ss}$ and relatively low spread. This indicates that the inclusion of all key physical processes including canopy processes, snow dynamics, soil storage, seepage, and baseflow contributes to balanced and robust representation of hydrological behavior.
- In contrast, the ablated models show varying degrees of performance degradation, highlighting the importance of different processes in various hydroclimatic regimes.
- In general, the simplest architecture, $GCHM_2$ (with no canopy or baseflow-generating storage) exhibits relatively low median performance and large variability, particularly in the *snowmelt* dominated (black) and *arid* (orange) catchments, suggesting that excluding baseflow and deeper soil storage processes limits the ability to reproduce delayed flow dynamics and recession behavior. However, in *recent-rainfall* catchments (red and blue), where fast runoff processes dominate and the role of deeper storage is less critical, its performance remains somewhat comparable to the un-ablated model.
- The reintroduction of baseflow storage in $GCHM_3^{NC}$ leads to noticeable improvements over $GCHM_2$, indicating better representation of slow-flow components. However, its performance remains slightly below that of $GCHM_5$ in *arid* (orange) and *snowmelt* dominated (black) catchments. This suggests that while incorporating baseflow improves model behavior, additional processes are required to achieve optimal performance across all hydroclimatic regimes.

- As might be expected, excluding the snow state in $GCHM_3^{NS}$, results in substantial performance degradation in *snowmelt* dominated catchments (black), and generally across all five hydro-climatic regimes, with lower median values and larger spread. This clearly demonstrates the critical role of explicitly representing snow accumulation and melt processes in such regions. The impact of removing the snow component is also noticeable in the *arid* regime (orange; where some catchments have snow), but less severe, in the *recent-rainfall-East* catchments (blue).

[98] Noticeably, the purely data-driven models ($LSTM_5$ variants) show mixed performance.

- In the *recent-rainfall* (red and blue) systems, they achieve slightly better median performance than $GCHM_5$. However, across *historical-rainfall* (green), *arid* (orange) and *snowmelt* dominated (black) catchments – where physical constraints appear to provide important guidance to the model – they exhibit lower median values, indicating less robustness and consistency. Of course, this may reflect, in part, the "*catchment-specific*" method of training adopted here, due to which the LSTM models are unable to learn from regional/continental scale large-sample training.

[99] The *snowmelt* dominated catchments (black) show the strongest sensitivity to model structure ablation with (as might be expected) clear performance drops when snow processes are excluded. *Arid* catchments (orange) also show strong sensitivity, particularly to the inclusion of baseflow and soil–atmosphere interactions, leading to the improved performance of $GCHM_5$ compared to the simpler models. In contrast, the *recent-rainfall* (blue and red) catchments show relatively smaller sensitivity to model structure, although $GCHM_5$ still provides more consistent performance.

[100] Overall, this analysis indicates that including key physical processes in the $GCHM_5$ architecture leads to improved accuracy, stability, and robustness, while ablation of these processes results in performance degradation. Additionally, while locally-trained pure-data-driven models can achieve excellent performance, they lack the consistency and generalizability provided by the physically informed structure of $GCHM_5$.

#### 4.2.2. Impact of Model Architecture Ablation on Streamflow Hydrographs

[101] We next evaluated the effects of model architecture ablation on the ability to reproduce characteristics of the streamflow hydrographs. **Figure 11** shows natural-scale and log-scale hydrographs for the five hydro-climatically representative catchments, considering both wet (a) and dry (b) years. We notice the following:

##### *4.2.2.1. Recent-rainfall-dominated (West) catchment [14222500; top row]*

- Ablated models exhibit strong performance during wet conditions, capturing the timing and magnitude of peaks reasonably well.
- However, the canopy-and-lower-zone-ablated $GCHM_2$ (black hydrograph) exhibits faster recession and underestimation of low flows due to the absence of a baseflow component. These deficiencies become substantially more pronounced during dry years, where $GCHM_2$ fails to sustain inter-event flows, leading to near-zero discharge (in log-space).
- The full five-state $GCHM_5$ architecture (blue hydrograph) consistently provides the best representation across wet and dry conditions, while the canopy-ablated $GCHM_3^{NC}$ (green-dashed hydrograph) and snow-ablated $GCHM_3^{NS}$ (orange-dashed hydrograph) perform similarly, indicating a limited role of canopy and snow processes relative to groundwater storage in this regime.

##### *4.2.2.2. Snowmelt-dominated catchment [9035900; second row]*

- The full five-state $GCHM_5$ architecture (blue) accurately captures the seasonal hydrograph in both wet and dry years, including the timing and magnitude of melt-driven peaks.
- The canopy-and-lower-zone-ablated $GCHM_2$ (black), and the canopy-ablated $\mathrm{GCHM}_3^{NC}$ (green) perform comparably, confirming the secondary importance of representing baseflow, and canopy storage.

- As might be expected, the snow-ablated, $\mathrm{GCHM}_3^{NS}$ (orange) shows substantial errors in both years, failing to reproduce the snowmelt dynamics altogether.
- This catchment clearly highlights the importance of representing both snow and groundwater processes.

#### *4.2.2.3. Historical-rainfall-dominated catchment [3331500; third row]*

- Again, during wet conditions, all ablated models exhibit strong performance, although the full five-state $GCHM_5$ architecture (blue) provides the best reproduction across peaks and recessions. Meanwhile, the canopy-ablated $GCHM_3^{NC}$ (green-dashed) and the snow-ablated, $\mathrm{GCHM}_3^{NS}$ (orange) show only minor deviations, suggesting that neither canopy nor snow processes dominate system behavior.
- In dry years, however, differences become clearer, with canopy-and-lower-zone-ablated $GCHM_2$ (black) failing to maintain baseflow, while the canopy-ablated $\mathrm{GCHM}_3^{NC}$ (green) and the snow-ablated, $\mathrm{GCHM}_3^{NS}$ (orange) variants retain more realistic low-flow dynamics due to the inclusion of groundwater storage.
- Again, the full five-state $GCHM_5$ architecture (blue) demonstrates the most consistent performance across wet and dry conditions.

#### *4.2.2.4. Recent-rainfall-dominated (East) catchment [1667500; fourth row]*

- This hydroclimatic regime is characterized by frequent flashy events. Again, the full five-state $GCHM_5$ architecture (blue) accurately captures the variability and magnitude of peaks in both wet and dry years.
- During wet conditions, progressive model ablation increasingly shows smoothing of the peaks, with the simplest canopy-and-lower-zone-ablated $GCHM_2$ (black) exhibiting the largest attenuation.
- Further, in dry years, the absence of a baseflow tank in $GCHM_2$ (black) leads to pronounced underestimation of low flows and loss of inter-event persistence.
- The canopy-ablated $\mathrm{GCHM}_3^{NC}$ (green) and the snow-ablated, $\mathrm{GCHM}_3^{NS}$ (orange) variants perform better than $GCHM_2$ (black) due to groundwater contributions, though they still underestimate variability compared to full five-state $GCHM_5$ architecture (blue), indicating the importance of full process coupling.

#### *4.2.2.5. Arid catchment [13161500; fifth row]*

- The effects of ablation are most pronounced in the *arid* catchment, most clearly demonstrating the critical role of baseflow storage under water-limited conditions.
- During wet years, the full five-state $GCHM_5$ (blue) captures both episodic peaks and recession behavior most accurately, while the canopy-and-lower-zone-ablated $GCHM_2$ (black) shows rapid post-event decline.
- In dry years, when flows are predominantly sustained by storage, the canopy-and-lower-zone-ablated $GCHM_2$ (black) fails almost entirely to represent low-flow persistence, producing near-zero discharge for extended periods. In contrast, the canopy-ablated $\mathrm{GCHM}_3^{NC}$ (green) and the snow-ablated, $\mathrm{GCHM}_3^{NS}$ (orange) variants maintain some level of baseflow due to the groundwater tank, though still with reduced accuracy compared to $\mathrm{MA}_5$.

#### *4.2.2.6. Summary of process ablation studies*

[102] Overall, the ablation study reveals that while simplified model structures may perform adequately under high-flow (wet) conditions, their deficiencies become evident under dry conditions, particularly in representing flow persistence and recession dynamics. The presence of a baseflow tank in $GCHM_3$ and $GCHM_5$ is essential for maintaining realistic low flows across all regimes, while an adequate representation of snow dynamics is

critical for *snowmelt* dominated systems. The full five-state $GCHM_5$ architecture consistently provides the most robust and transferable performance across diverse hydroclimatic conditions and regimes.

### 4.2.3. Impact of Model Architecture Ablation on Flow Duration Curves (FDCs)

[103] We similarly evaluated the impact of model architecture (process) ablation on the flow duration curves (FDCs; see **Figure 12**). Panel (a) shows the high-flow segment (exceedance probability, EP = 0-5%), while panel (b) presents the full FDC (EP = 0-100%), and panel (c) focuses on the low-flow segment (EP = 80-100%), allowing a detailed assessment of model behavior across the entire flow regime.

#### *4.2.3.1. High Flows*

[104] For high flows (panel a), all models generally reproduce the overall shape of the observed FDCs (red dashed line) across catchments, indicating their ability to capture the upper tail behavior. However, systematic differences are evident.

- The full $GCHM_5$ model (blue) consistently provides the closest agreement with observations, particularly in the flashy *recent-rainfall* catchments (14222500 and 1667500), where it better captures the steepness and curvature of the high-flow tail.
- The ablated models show progressive deviation: $GCHM_3^{NC}$ (green) performs comparably to $GCHM_5$ in most cases, suggesting a limited role of canopy storage for extreme flows, whereas $GCHM_3^{NS}$ (orange) significantly underestimates high flows in the *snowmelt* dominated catchment (9035900), reflecting the absence of snow processes.
- $GCHM_2$ (black) shows mixed behavior, often underestimating the highest flows or producing overly steep declines, indicating limitations in representing rapid runoff generation with an overly-simplified structure.

#### *4.2.3.2. Entire FDC*

[105] Across the full FDCs (panel b), differences among model structures become more pronounced, particularly in the mid- and low-flow ranges.

- The full $GCHM_5$ model (blue) consistently tracks the observed curves across all exceedance probabilities, indicating a balanced representation of both fast- and slow-flow components.
- $GCHM_3^{NC}$ (green) and $GCHM_3^{NS}$ (orange) generally follow similar trends but exhibit biases depending on the dominant processes: $GCHM_3^{NS}$ deviates substantially in the *snowmelt* dominated catchment, while $GCHM_3^{NC}$ shows minor differences in *rainfall-driven* systems.
- In contrast, $GCHM_2$ systematically departs from observations in the lower half of the FDC, with steeper slopes and earlier drop-offs, reflecting its inability to sustain flows due to the absence of a baseflow storage component.

#### *4.2.3.3. Low Flows*

[106] The low-flow segment (panel c) highlights the most significant structural differences among models.

- The full $GCHM_5$ model (blue) provides the best agreement with observed low flows across all catchments, maintaining realistic curvature and magnitude in the high exceedance range.
- $GCHM_3^{NC}$ (green) and $GCHM_3^{NS}$ (orange) also perform reasonably well, benefiting from the inclusion of a groundwater/baseflow tank that supports flow persistence. However, $\mathrm{GCHM}_3^{NS}$ shows additional bias in *snow-influenced* catchments, where missing snow dynamics propagate into low-flow errors.
- In contrast, $GCHM_2$ (black) exhibits pronounced underestimation of low flows across all regimes, with a sharp downturn in the tail of the FDC. This behavior is particularly evident in *arid* (13161500) and rainfall-driven catchments, where sustained baseflow dominates streamflow during dry periods. The absence of a baseflow reservoir in $GCHM_2$ leads to rapid depletion of storage and an inability to reproduce observed flow persistence.

## 5. Conclusions

[107] This study has reported on the development and evaluation of a relatively-parsimonious, physically-conceptually-interpretable, mass-conserving Gated Catchment Hydrology Model (GCHM) that integrates data-driven learning with explicit conceptual representation of hydrologic storage, fluxes, and process controls. Across contrasting hydroclimatic regimes, we demonstrated that a relatively simple five-state representation can reproduce important streamflow characteristics while maintaining interpretable internal states and physically constrained flux pathways. The experiments showed that model performance and process representation depend strongly on hydroclimatic conditions, with groundwater storage, snow processes, and evapotranspiration formulation playing particularly important roles in determining model robustness.

[108] The principal conclusions of this study are:

- A parsimonious, mass-conserving GCHM-type architecture can effectively combine machine-learning flexibility with physically interpretable hydrologic structure. Explicit representation of canopy interception, snow storage, upper-soil water, groundwater storage, and runoff generation provides direct correspondence between model states, fluxes, and hydrologic processes while allowing key process relationships to be learned from data.
- The GCHM can reproduce daily streamflow dynamics across diverse hydroclimatic conditions. However, its performance is regime dependent, with *snowmelt* dominated and *historical-rainfall* catchments exhibiting the best and most stable performance, and arid regimes presenting the greatest challenge (particularly for rapid peak flows and low-flow conditions) although remarkably good performance was observed for several arid catchments. Overall, long-term mass balance was consistently well represented across regimes.
- The learned gating functions exhibit hydrologically plausible behavior and provide a direct means of interpreting the model's learned flux-process relationships. Consistent with understanding, snowmelt activation and rainfall–snowfall partitioning responds systematically to temperature, while infiltration, saturation excess, seepage, and groundwater outflow vary consistently with antecedent storage conditions and hydroclimatic regime.
- Explicit representation of multiple runoff pathways enables the model to distinguish fast and slow components of streamflow generation. In rainfall-responsive catchments, infiltration- and saturation-excess processes contribute strongly during high-flow events, whereas upper-soil outflow and groundwater baseflow provide more persistent contributions. Baseflow becomes particularly important during dry periods by sustaining streamflow in the absence of substantial precipitation input.
- Groundwater storage is a critical architectural component for reproducing delayed flow dynamics and low-flow behavior. Ablation of the lower storage and associated baseflow pathway led to unrealistically rapid recession and a reduced ability to sustain inter-event and dry-period flows, demonstrating that deeper storage provides an essential memory component in catchments where slow hydrologic pathways are important.
- Explicit representation of snow storage is essential in snowmelt-influenced catchments. Removing the snow state produced substantial performance degradation in *snowmelt* dominated systems, confirming that snow accumulation and melt must be explicitly represented when seasonal snow storage exerts strong control over streamflow timing and magnitude.
- The model generates realistic snow dynamics even without direct training to observed SWE data. Comparison with the independent UA-SWE product shows that the model reproduces the broad seasonal evolution and timing of snow accumulation and melt, indicating that the learned streamflow-constrained snow states contain meaningful hydrologic information. Remaining discrepancies, particularly in arid catchments, highlight the influence of phase-partitioning uncertainty, melt thresholds, and spatial heterogeneity.

- Evapotranspiration formulations substantially influence model robustness, particularly in arid systems. The conventional Penman–Monteith formulation resulted in weaker and more variable performance, whereas greater flexibility in representing surface resistance and meteorological controls improved model performance. The largest gains occurred in arid and *historical-rainfall* catchments, demonstrating the sensitivity of water-limited systems to ET representation.
- Systematic ablation confirms that explicit process representation improves the consistency of model behavior across hydroclimatic regimes. Overall, while simplified model structures may perform adequately under high-flow (wet) conditions, their deficiencies become evident under dry conditions, particularly in representing flow persistence and recession dynamics. So, although simple architectures and purely data-driven LSTM models can achieve competitive performance in some *rainfall*-dominated catchments, they show greater sensitivity and variability in *snowmelt* dominated and *arid* systems. The full GCHM provides a more balanced representation by retaining the complementary roles of snow storage, soil storage, groundwater, and multiple runoff pathways.

[109] Overall, our results demonstrate that physical constraints and machine learning can be integrated at the architectural level rather than treated as separate modeling paradigms. By embedding mass conservation directly into the computational units and allowing process controls to be learned through context-dependent gates, the GCHM provides a framework in which predictive performance, physical consistency, and process interpretability can be addressed simultaneously.

[110] The broader implication is that hydrologic machine-learning models need not be either fully process-based or fully black-box. A carefully designed, parsimonious architecture can retain explicit water-storage and flow relationships while learning nonlinear process behavior from observations, thereby providing a practical pathway toward hydrologic models that support both prediction and scientific process understanding.

[111] Ongoing work is exploring the sensitivity of models based in the GCHM framework to the informativeness and quality of contextually-relevant data sources used to drive the model, and to inform the learned flux-process related gating functions; these may help to shed light on the reasons for the reported poor performance of contemporary physical-conceptual and machine-learning-based models in certain catchments and/or hydro-climatic regimes.

## Acknowledgments

Funding for this project was provided by the National Oceanic and Atmospheric Administration (NOAA), awarded to the Cooperative Institute for Research on Hydrology (CIROH) through the NOAA Cooperative Agreement with The University of Alabama, NA22NWS4320003/A25-0353-S006. We gratefully acknowledge *Peter A. Troch* for generously providing access to his HPC account, enabling the computational analyses presented in this work. The second author (HVG) acknowledges support for a 4-month research visit to the Karlsruhe Institute of Technology, Germany, provided by the KIT International Excellence Fellowship Award program, and is grateful for the inspiration and encouragement provided by members of the Information Theory in the Geosciences group (geoinfotheory.org), and by *Uwe Ehret*, *Ralf Loritz*, *Erwin Zehe*, *Anneli Guthke*, *Manuel Alvarez-Chavez* and several others during his visits to Germany over the past several years.

## Tables

**Table 1.** Alternative formulations ($S_1$ to $S_4$) of the evapotranspiration gating function $G_t$ tested in the $GCHM_5$ architecture, where the ET loss is $L_t = G_t \cdot X_t$. Here σ is the sigmoid function; $X_t$ is the state variable (canopy, soil-surface or lower-soil storage); $PM_t^{Trad}$ and $PM_t^{NN}$ are Penman–Monteith ET estimates using traditional empirical resistances and neural-network surface resistance, respectively; $DSW_t$, $VPD_t$, $T_t$, $\Delta_t$ and $u_t$ are downwelling shortwave radiation, vapor pressure deficit, air temperature, slope of the saturation vapor pressure curve and wind speed; $FNN_t$ is the output of a feedforward neural network; and a–g are fitted coefficients.

| Formulation | Equation of Gate | Architecture Identifier |
|---|---|---|
| $S_1$ | $\sigma\left(a \cdot X_t + b \cdot PM_t^{Trad} + c\right)$ | $GCHM_5^{PM:Trad}$ |
| $S_2$ | $\sigma(a \cdot X_t + b \cdot PM_t^{NN} + c)$ | $GCHM_5^{PM:NN}$ |
| $S_3$ | $\sigma(a \cdot X_t + b \cdot DSW_t + c \cdot VPD_t + d \cdot T_t + e \cdot \Delta_t + f \cdot u_t + g)$ | $GCHM_5^{Sigmoid}$ |
| $S_4$ | $\sigma(a \cdot X_t + b \cdot FNN_t + c)$ | $GCHM_5^{NN}$ |

## Figures

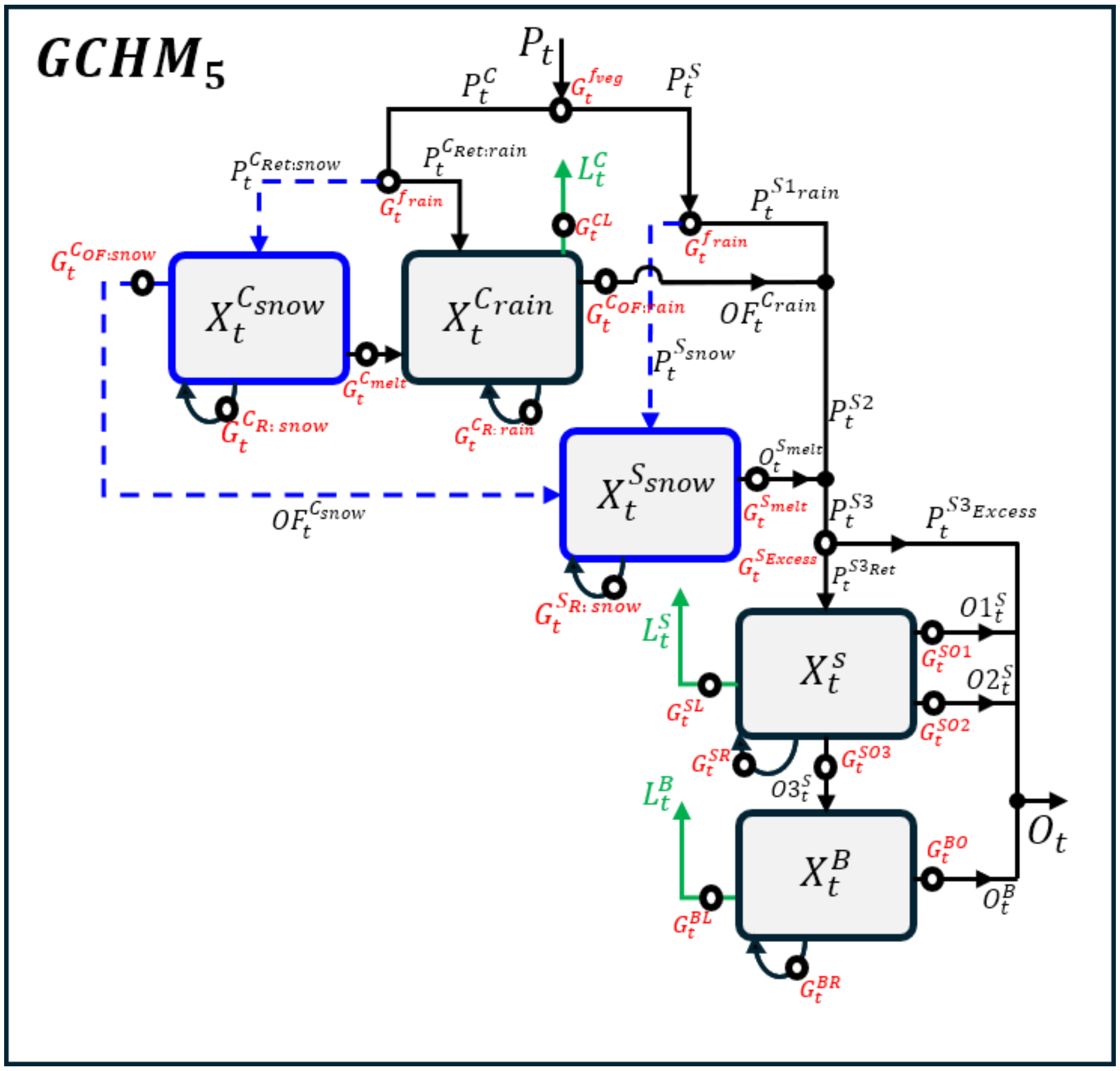

**Figure 1**: Architecture of the five-state, mass-conserving Gated Catchment Hydrology Model ($GCHM_5$). The five states are canopy snow and rain interception storage, ground snowpack, upper-soil storage and lower-soil (baseflow) storage. They are connected by learned, state-dependent gated fluxes, and the outflow pathways (infiltration excess, saturation excess, soil outflow and baseflow) sum to total streamflow.

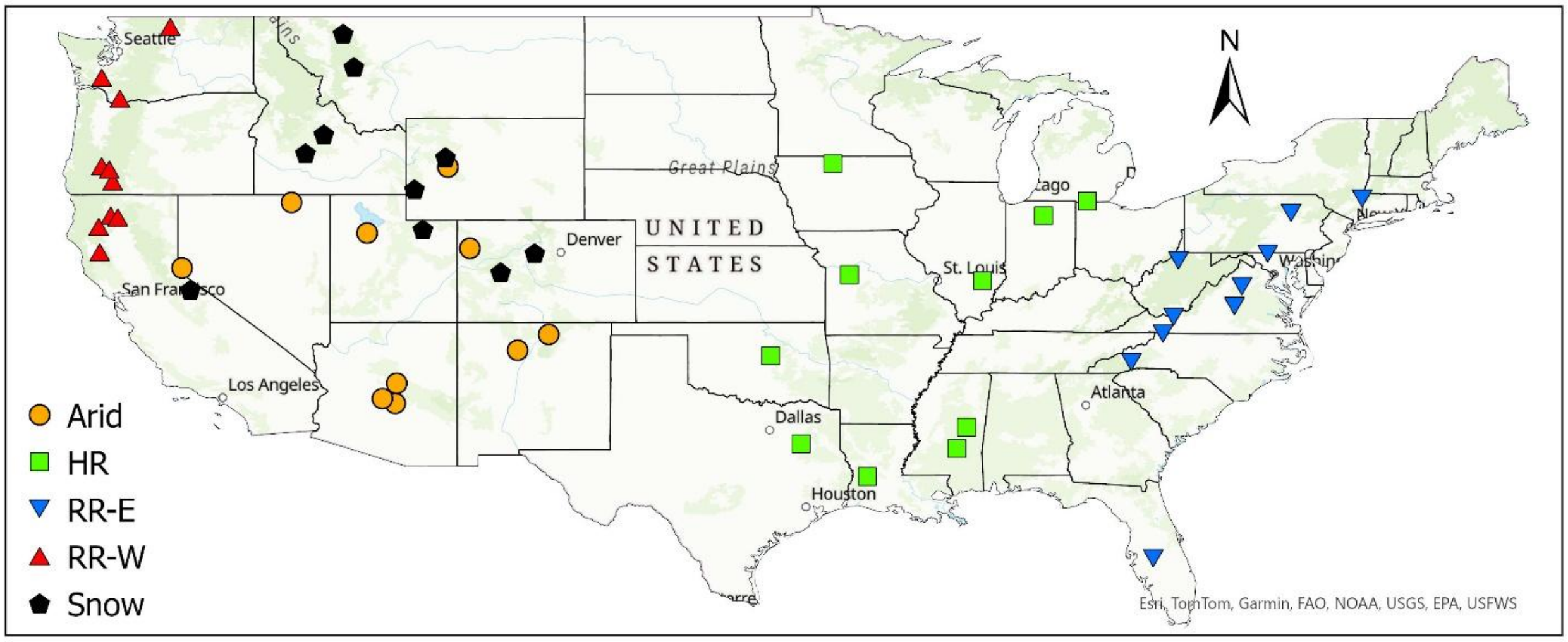

**Figure 2.** Locations of the 50 CAMELS study catchments, classified by dominant streamflow-response regime: arid (Arid), historical-rainfall-dominated (HR), recent-rainfall-dominated East (RR-E) and West (RR-W), and snowmelt-dominated (Snow). Catchment attributes are listed in Table S1.

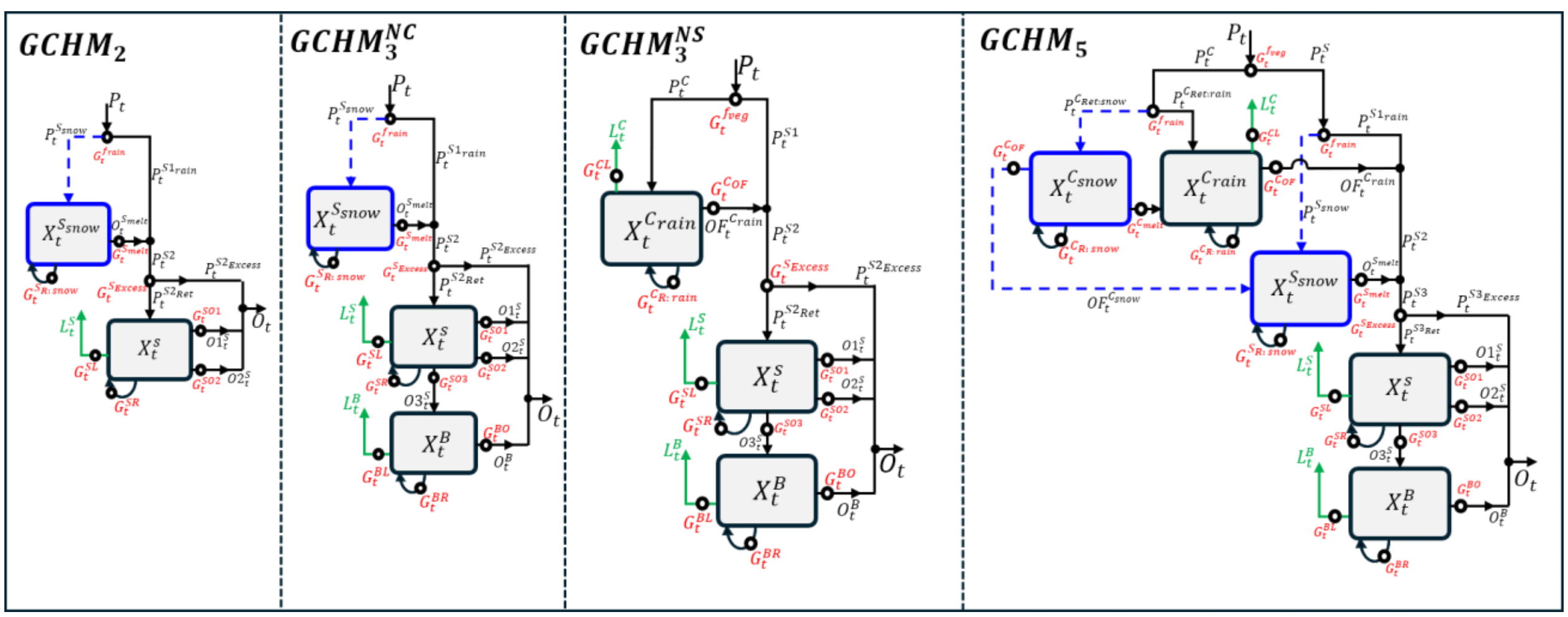

**Figure 3.** The full five-state architecture ($GCHM_5$) and the three ablated versions used to assess the importance of individual processes: $GCHM_3^{NC}$ (no canopy storage), $GCHM_3^{NS}$ (no snow storage) and $GCHM_2$ (no canopy or baseflow storage; snow and soil storage only).

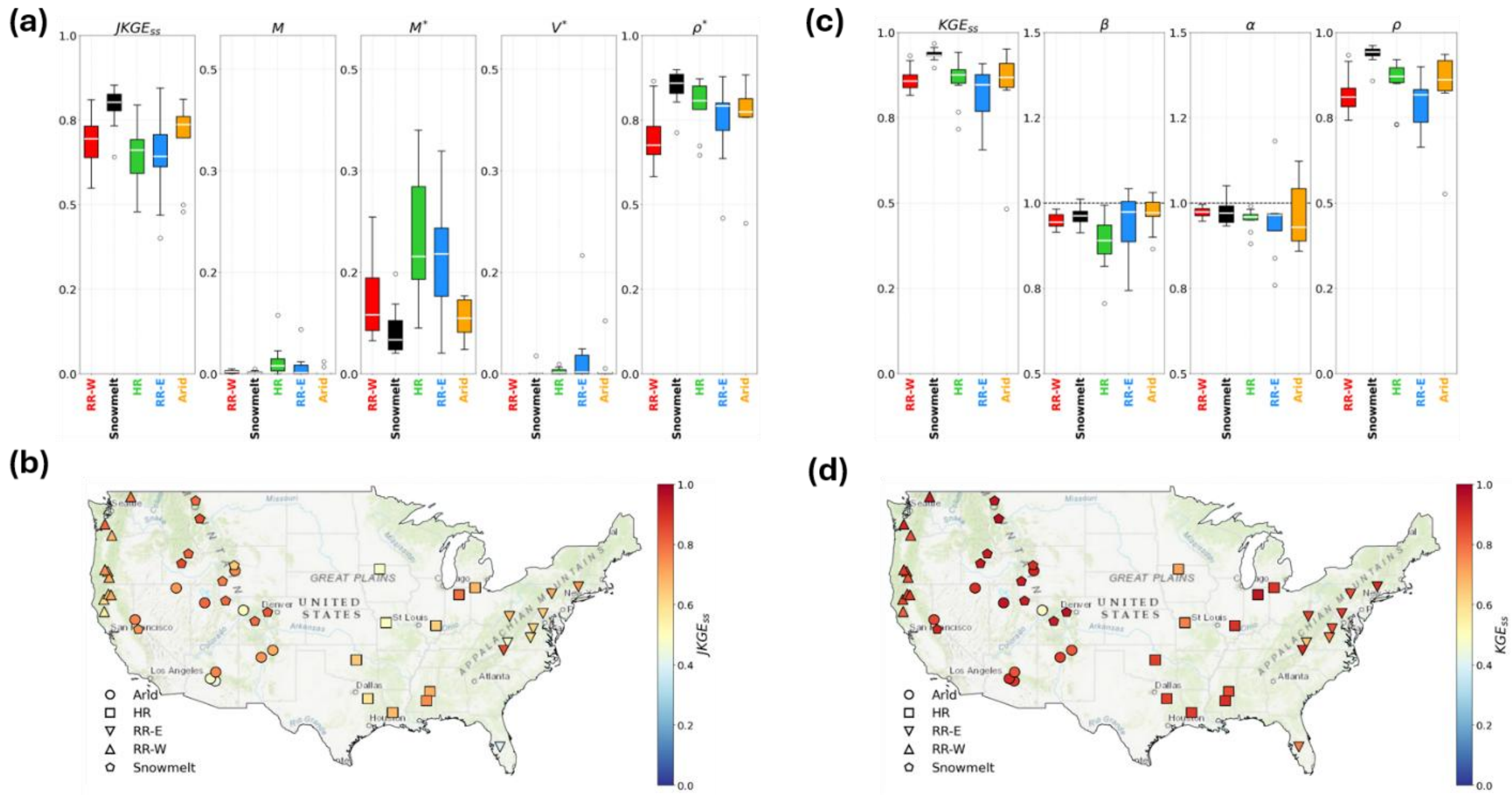

**Figure 4.** Performance of $GCHM_5$ evaluated over the complete record of each catchment (one value per catchment), by hydroclimatic regime. (a) Box plots of the non-stationary skill score $JKGE_{ss}$ and its components M, $M^*$, $V^*$ and $\rho^*$; (b) map of catchment-level $JKGE_{ss}$; (c) box plots of the stationary skill score $KGE_{ss}$ and its components β, α and ρ; (d) map of catchment-level $KGE_{ss}$. A value of 1 is perfect for $JKGE_{ss}$, $KGE_{ss}$, $\rho^*$ and ρ; 0 is perfect for the error terms M, $M^*$ and $V^*$; and 1 is ideal for β and α. Boxes show the median and interquartile range, and circles are outliers. Marker shapes in (b) and (d) denote regime.

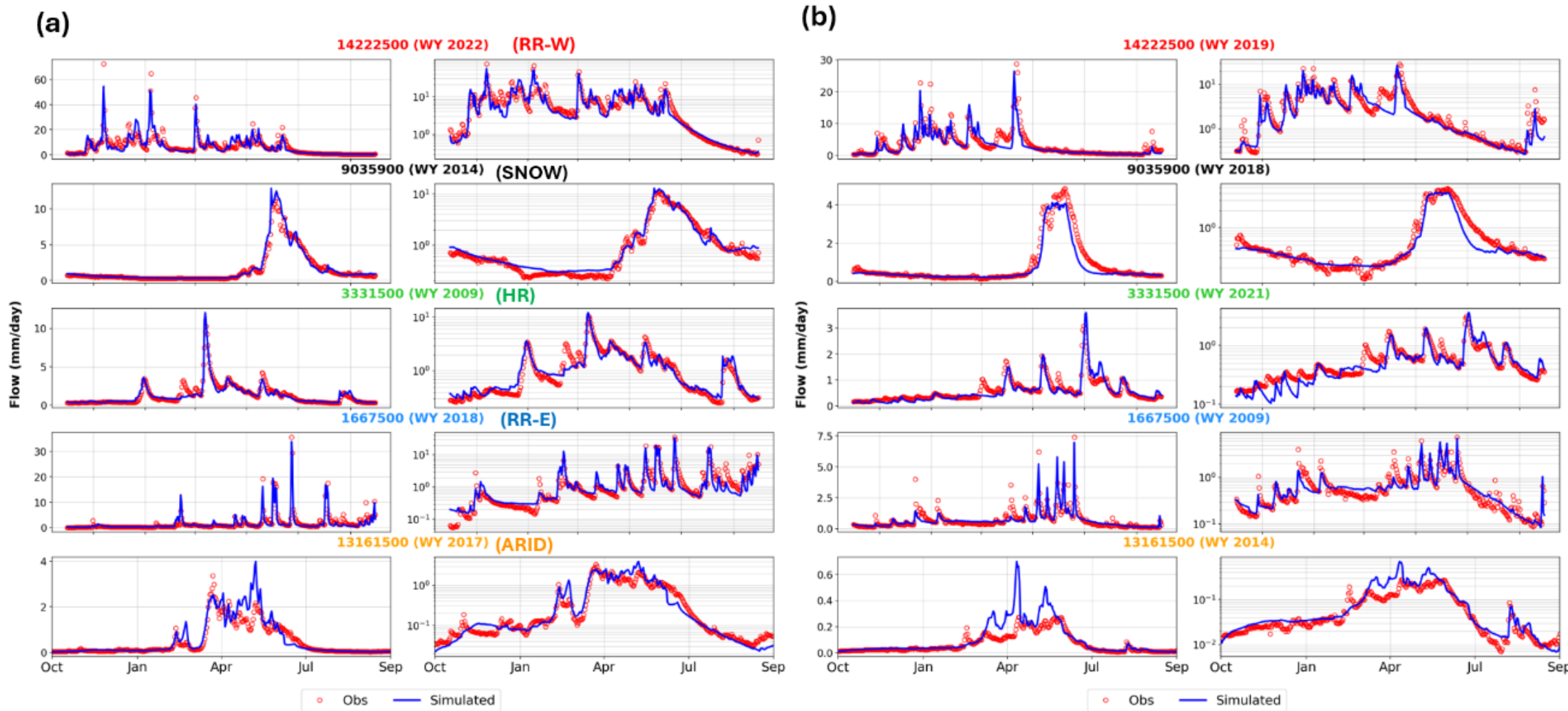

**Figure 5.** Observed and simulated streamflow hydrographs from $GCHM_5$ for five representative catchments, one per hydroclimatic regime (rows, top to bottom: RR-W, Snow, HR, RR-E and Arid), for (a) a wet year and (b) a dry year.

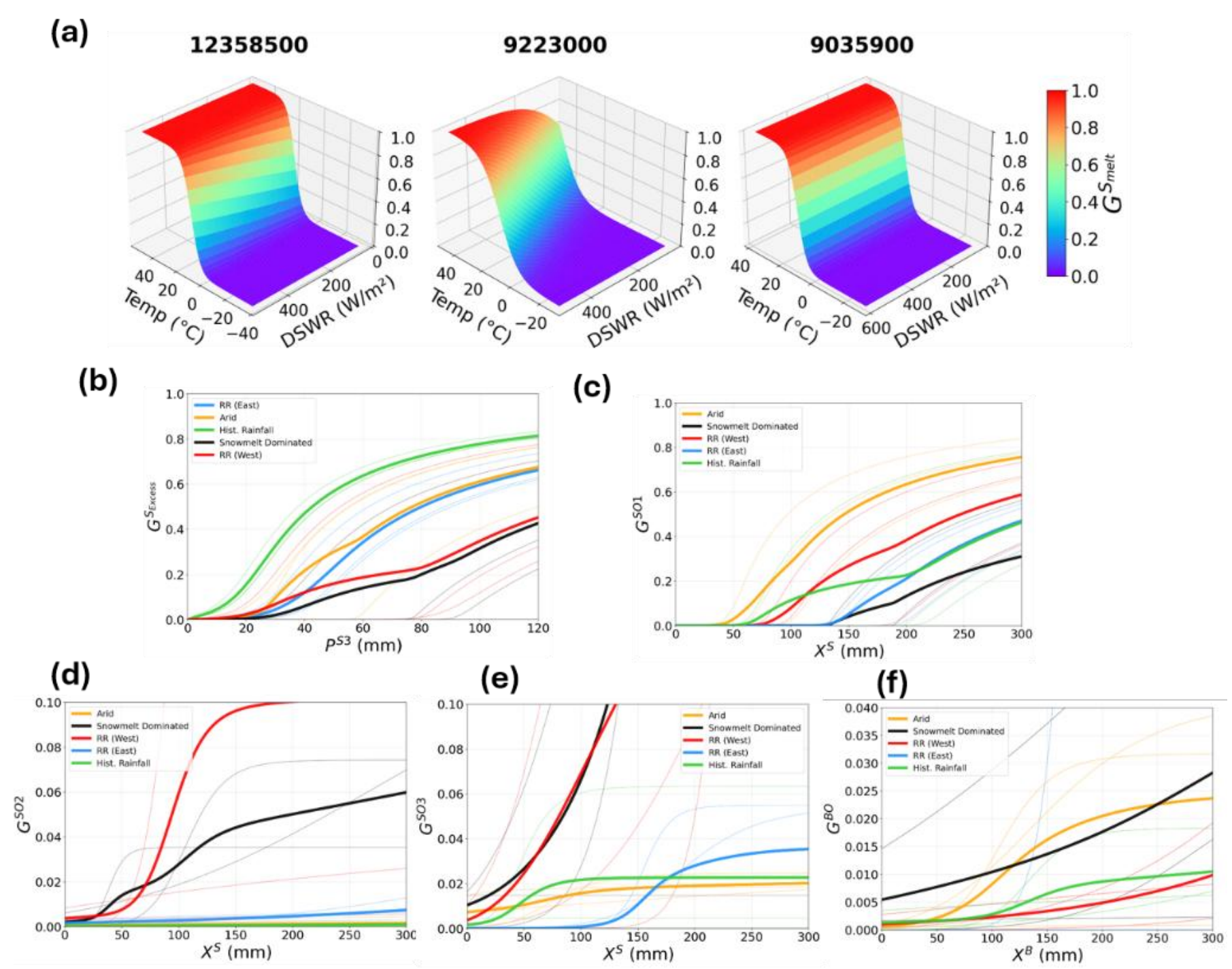

**Figure 6.** Learned activation gating functions of $GCHM_5$. (a) Snowmelt gate $G^{S_melt}$ as a function of air temperature (Temp) and downwelling shortwave radiation (DSWR) for the three snowmelt-dominated catchments shown (USGS 12358500, 09223000 and 09035900); (b) infiltration-excess gate versus $P^{S3}$; (c) saturation-excess gate $G^{SO1}$ versus upper-soil storage $X^S$; (d) soil-outflow gate $G^{SO2}$ versus $X^S$; (e) seepage gate $G^{SO3}$ versus $X^S$; (f) baseflow-outflow gate $G^{BO}$ versus lower-zone storage $X^B$. In (b)–(f), thin lines are individual catchments and thick lines are the mean for each hydroclimatic regime.

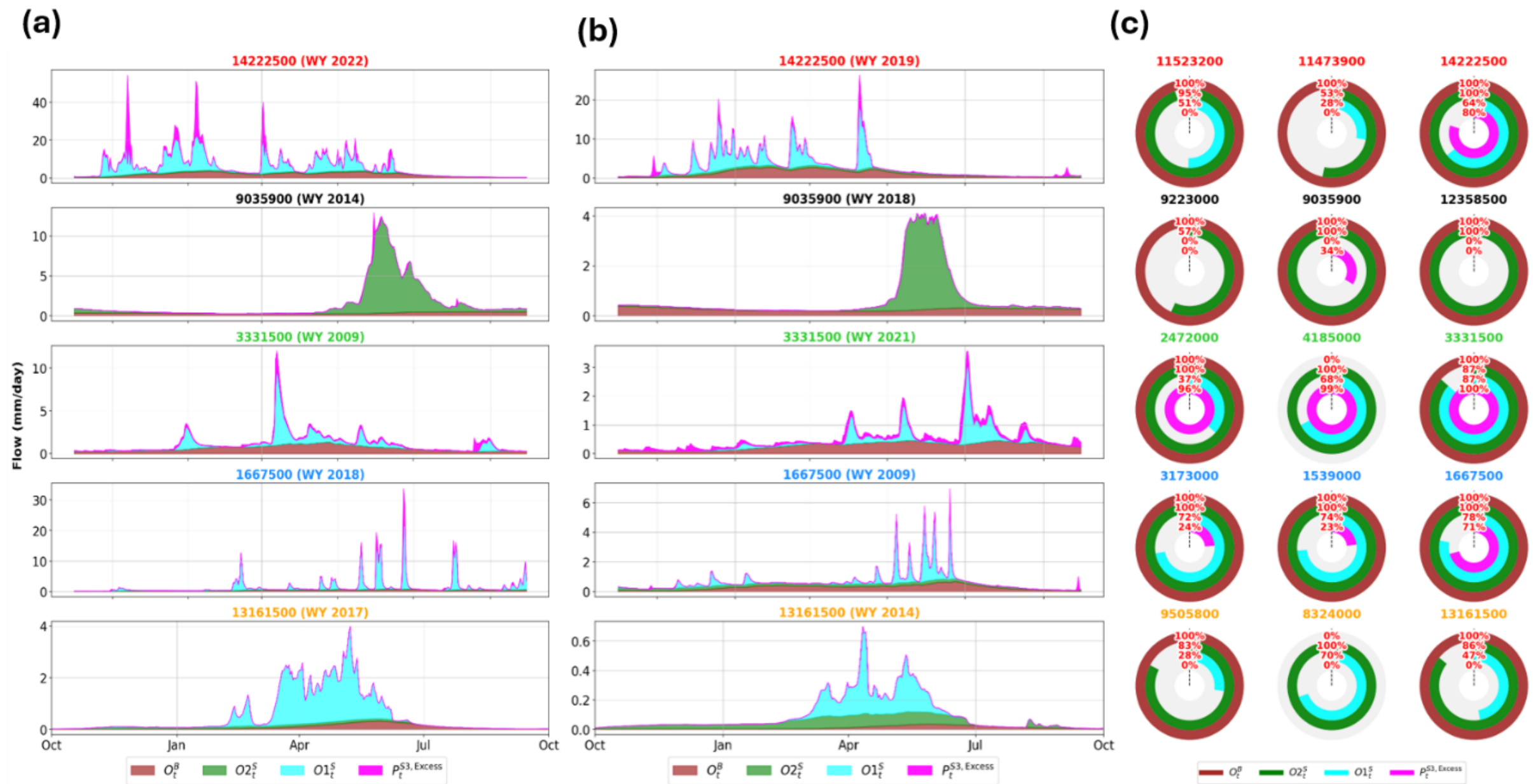

**Figure 7.** Contributions of the four flow pathways to simulated streamflow for five representative catchments (one per regime) in (a) a wet year and (b) a dry year, and (c) the frequency of activation of each pathway, expressed as the percentage of time steps (over 20 years) in which the pathway flux exceeds 0.01 mm.

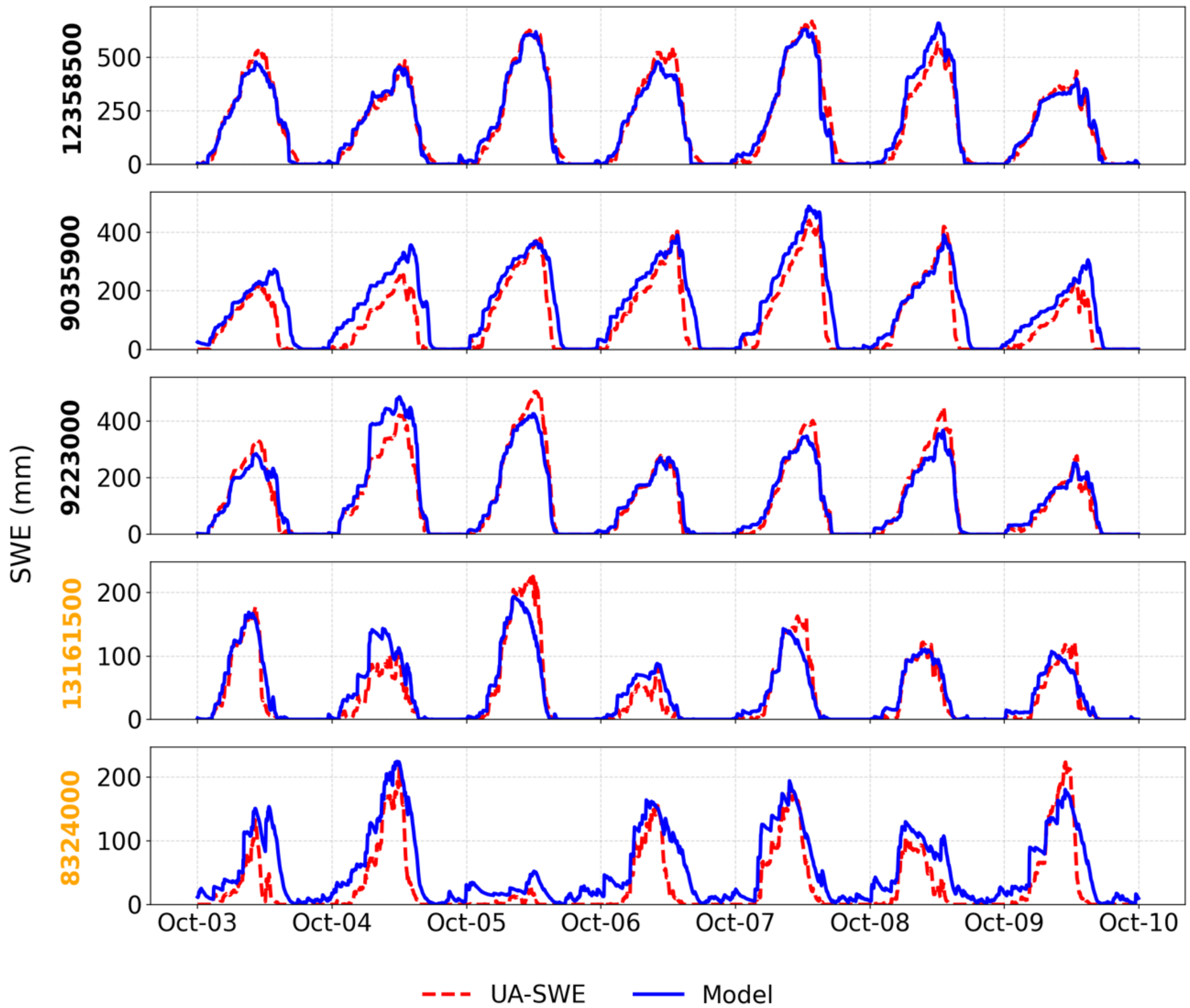

**Figure 8.** Model-simulated snow water equivalent (SWE; Model) compared with the independent UA-SWE product for the five representative catchments (USGS gauge IDs at left) for water years 2004 to 2010. The model was trained on streamflow only and SWE observations were not used in training.

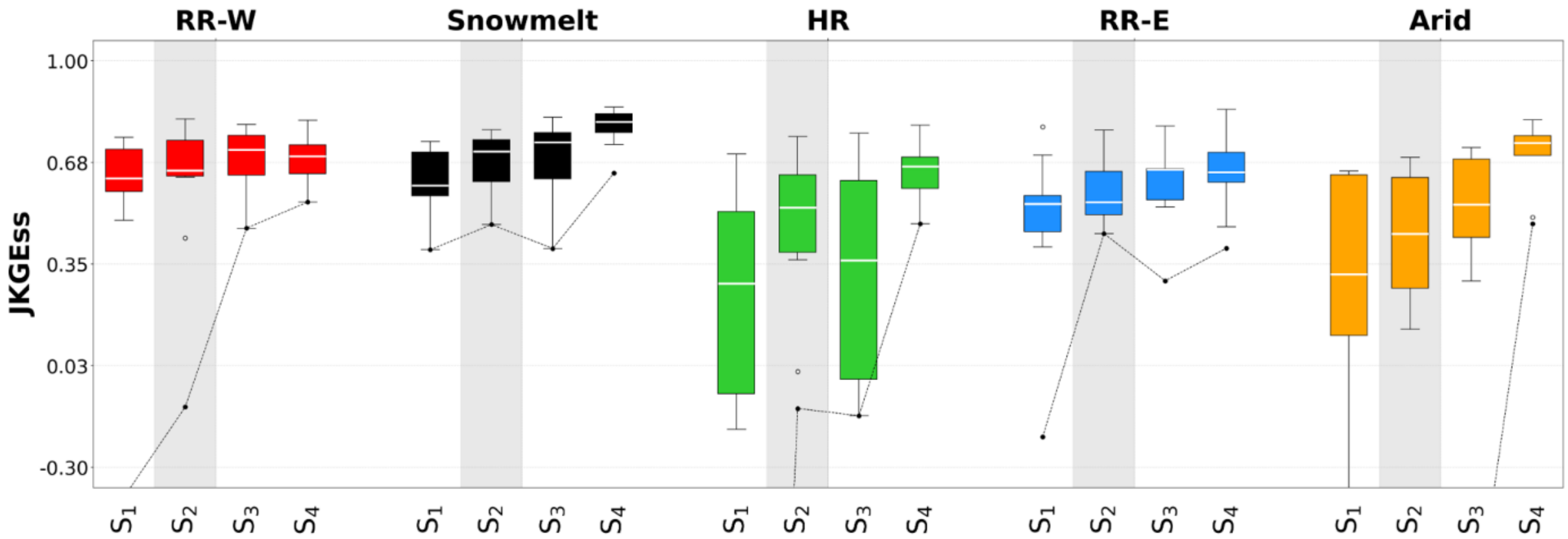

**Figure 9.** Effect of the evapotranspiration gating formulation ($S_1$ to $S_4$; Table 1) on model performance. Box plots show $JKGE_{ss}$ across the catchments of each hydroclimatic regime (RR-W, Snow, HR, RR-E and Arid).

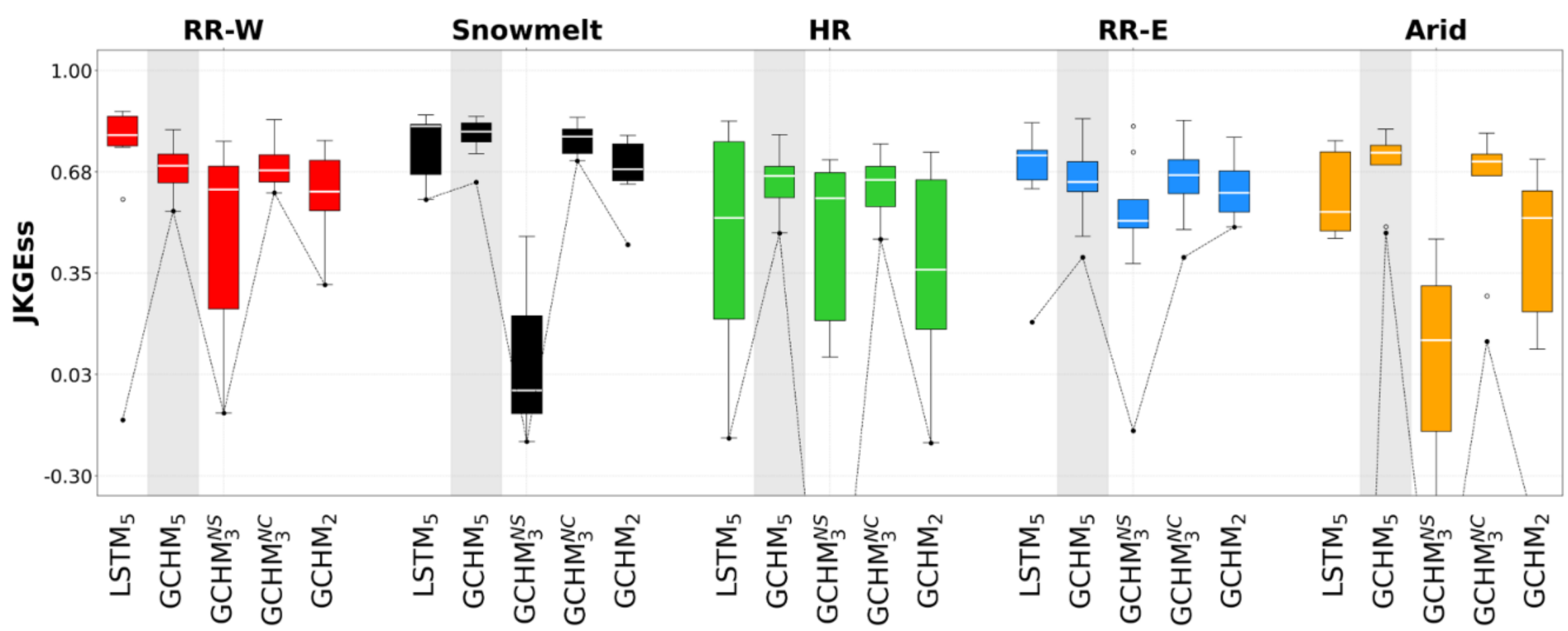

**Figure 10.** Comparison of $JKGE_{ss}$ for the full $GCHM_5$ model, its ablated versions ($GCHM_3^{NS}$, $GCHM_3^{NC}$ and $GCHM_2$) and the purely data-driven $LSTM_5$ model, by hydroclimatic regime. Boxes show the median and interquartile range across catchments; whiskers extend to 1.5 times the interquartile range and circles are outliers. The shaded band marks the full $GCHM_5$ model in each regime.

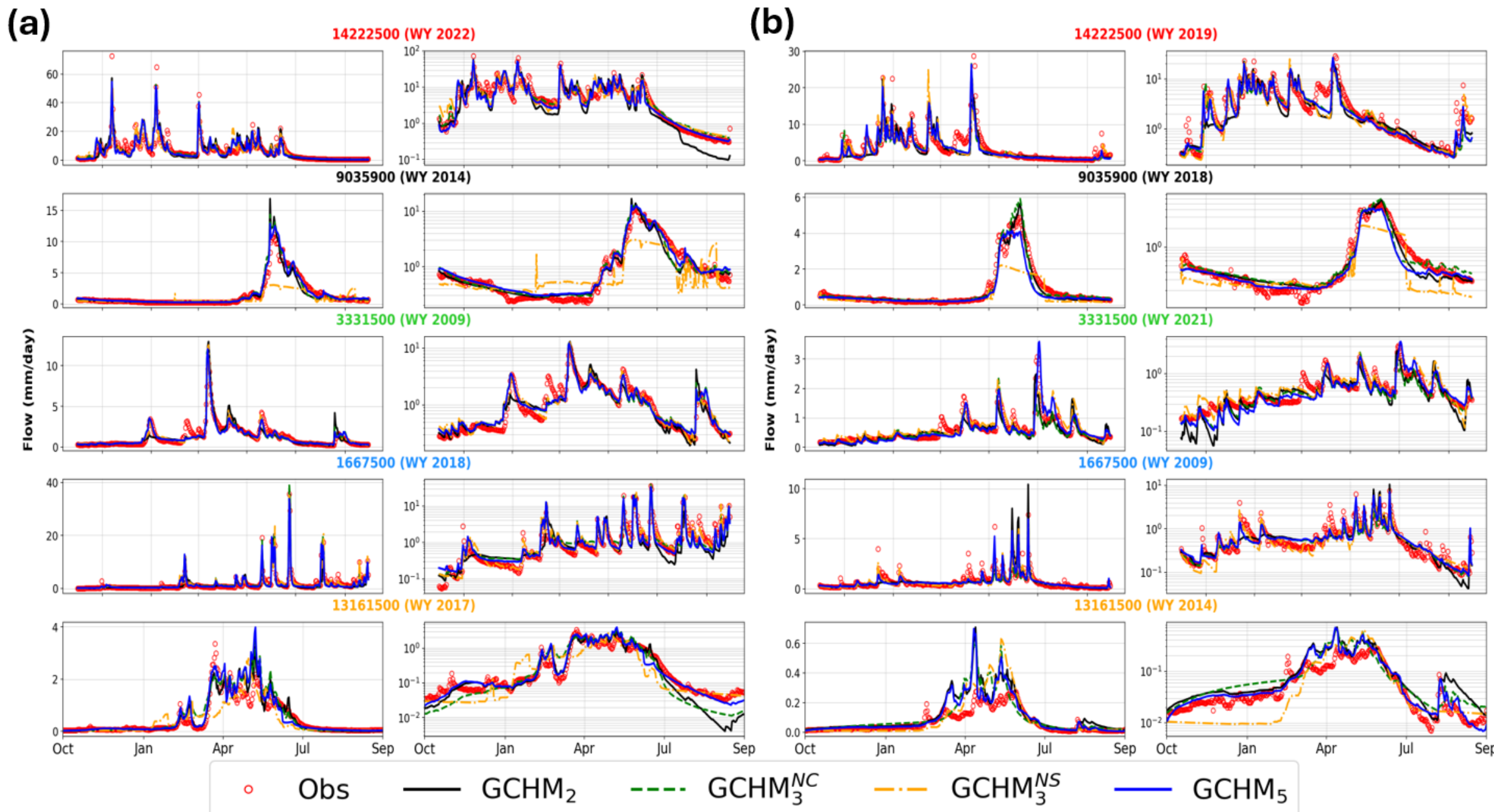

**Figure 11.** Observed and simulated hydrographs from the full GCHM$_5$ model and its ablated versions on natural and logarithmic flow scales for the five representative catchments (one per regime) in (a) a wet year and (b) a dry year.

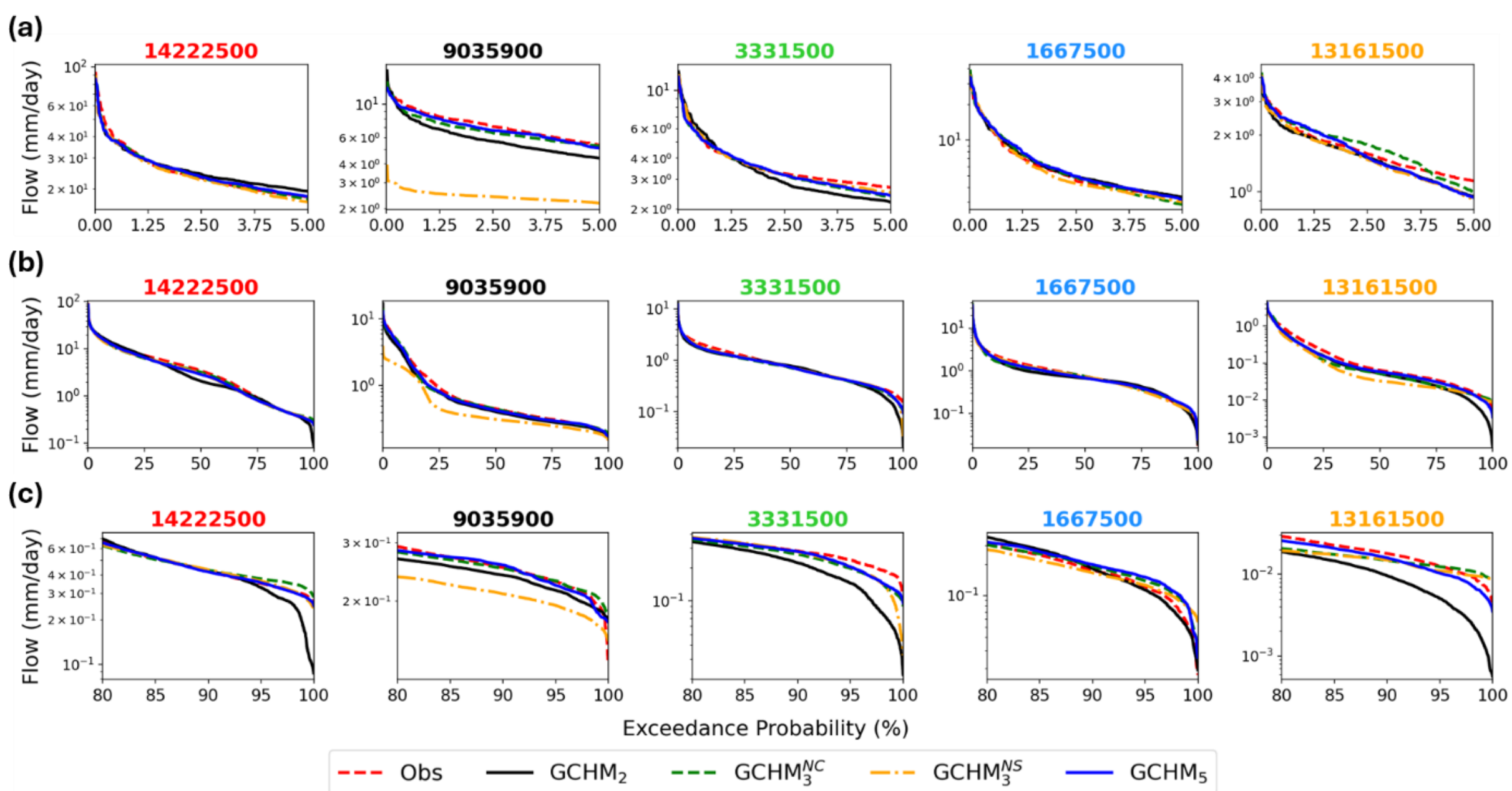

**Figure 12.** Effect of architecture ablation on flow duration curves (FDCs) for the representative catchments: (a) high-flow segment (exceedance probability, EP = 0–5%); (b) full FDC (EP = 0–100%); (c) low-flow segment (EP = 80–100%). The observed FDC is shown as a red dashed line.

## Supplementary Materials

M Jawad[1], HV Gupta[1], MA Farmani[1], A Behrangi[1], and GY Niu[1]

[1]Department of Hydrology and Atmospheric Sciences, The University of Arizona, Tucson, AZ, USA, 85721.

**Table S1.** The 50 CAMELS catchments used in this study, grouped by hydroclimatic regime (ten catchments per regime): USGS gauge ID, gauge name, latitude and longitude (decimal degrees), drainage area and regime classification.

| ID | Name | Latitude | Longitude | Area (km²) | Criteria |
|---|---|---|---|---|---|
| 11473900 | Middle Fork Eel River, near Dos Rios, CA | 39.79 | -123.07 | 1930.44 | Recent rainfall-dominated (West) |
| 11523200 | Trinity River above Coffee Creek, CA | 41.23 | -122.64 | 385.09 | |
| 14222500 | East Fork Lewis River, near Heisson, WA | 45.81 | -122.29 | 329.06 | |
| 14321000 | Umpqua River near Elkton, OR | 43.16 | -123 | 9452.17 | |
| 14359000 | Rogue River at Raygold near Central Point, OR | 42.56 | -122.57 | 5526.27 | |
| 14308000 | South Umpqua River at Tiller, OR | 43.03 | -122.7 | 1159.87 | |
| 11342000 | Sacramento River at Delta, CA | 41.16 | -122.37 | 1095.02 | |
| 12027500 | Chehalis River near Grand Mound, WA | 46.63 | -123 | 2296.5 | |
| 11530000 | Trinity River at Hoopa, CA | 40.79 | -123.11 | 7378.78 | |
| 12449500 | Methow River at Twisp, WA | 48.62 | -120.3 | 3365.92 | |
| 09035900 | South Fork of Williams Fork near Leal, CO | 39.73 | -106.01 | 71.67 | Snowmelt Dominated |
| 09223000 | Hams Fork below Pole Creek, near Frontier, WY | 42.23 | -110.72 | 333.86 | |
| 12358500 | Middle Fork Flathead River near West Glacier, MT | 48.33 | -113.53 | 2917.03 | |
| 09292500 | Yellowstone River near Altonah, UT | 40.66 | -110.39 | 339.08 | |
| 13302500 | Salmon River at Salmon, ID | 44.39 | -114.28 | 9734.4 | |
| 13186000 | South Fork Boise River near Featherville, ID | 43.65 | -115 | 1655.3 | |
| 10296000 | West Walker River below Little Walker River, near Coleville, CA | 38.29 | -119.52 | 462.4 | |
| 12340000 | Blackfoot River near Bonner, MT | 47.02 | -113.1 | 5926.09 | |
| 06225500 | Wind River near Crowheart, WY | 43.49 | -109.51 | 4994.84 | |
| 09132500 | North Fork Gunnison River near Somerset, CO | 38.98 | -107.34 | 1361.42 | |
| 02472000 | Leaf River near Collins, MS | 32.04 | -89.45 | 1927.71 | Historical rainfall-dominated |
| 04185000 | Tiffin River at Stryker, OH | 41.74 | -84.34 | 1064.24 | |
| 03381500 | Little Wabash River at Carmi, IL | 38.63 | -88.44 | 8044.15 | |
| 08015500 | Calcasieu River near Kinder, LA | 30.95 | -92.96 | 4473.15 | |
| 05479000 | East Fork Des Moines River at Dakota City, IA | 43.22 | -94.31 | 3373.18 | |
| 08032000 | Neches River near Neches, TX | 32.23 | -95.56 | 2940.92 | |
| 03331500 | Tippecanoe River near Ora, IN | 41.17 | -86.04 | 2248.64 | |
| 02482000 | Pearl River at Edinburg, MS | 32.88 | -89.07 | 2342.99 | |
| 06908000 | Blackwater River at Blue Lick, MO | 38.86 | -93.67 | 2923.94 | |
| 07243500 | Deep Fork near Beggs, OK | 35.68 | -96.76 | 5210.24 | |
| 02296750 | Peace River at SR 70 at Arcadia, FL | 27.7 | -81.75 | 3624.47 | Recent rainfall-dominated (East) |
| 01539000 | Fishing Creek near Bloomsburg, PA | 41.24 | -76.34 | 705.23 | |
| 01667500 | Rapidan River near Culpeper, VA | 38.36 | -78.27 | 1210.77 | |
| 03164000 | New River near Galax, VA | 36.5 | -81.36 | 2963.75 | |
| 03173000 | Walker Creek at Bane, VA | 37.16 | -80.95 | 775.86 | |
| 01643000 | Monocacy River at Jug Bridge near Frederick, MD | 39.63 | -77.26 | 2122.46 | |
| 03114500 | Middle Island Creek at Little, WV | 39.37 | -80.76 | 1196.06 | |
| 07069500 | Spring River at Imboden, AR | 36.46 | -91.6 | 3033.91 | |
| 01200000 | Tenmile River near Gaylordsville, CT | 41.8 | -73.55 | 517.39 | |
| 02030500 | Slate River near Arvonia, VA | 37.58 | -78.56 | 569.39 | |
| 06224000 | Bull Lake Creek above Bull Lake, WY | 43.09 | -109.42 | 485.7 | Arid Catchment |
| 08271000 | Rio Lucero near Arroyo Seco, NM | 36.52 | -105.46 | 46.39 | |
| 08324000 | Jemez River near Jemez, NM | 35.89 | -106.67 | 1222.22 | |
| 09306242 | Corral Gulch near Rangely, CO | 39.9 | -108.55 | 87.71 | |
| 09505800 | West Clear Creek near Camp Verde, AZ | 34.6 | -111.43 | 624.73 | |
| 10172800 | South Willow Creek near Grantsville, UT | 40.5 | -112.58 | 76.51 | |
| 10310500 | Clear Creek near Carson City, NV | 39.12 | -119.85 | 47.65 | |
| 13161500 | Bruneau River at Rowland, NV | 41.7 | -115.56 | 990.36 | |
| 09513780 | New River near Rock Springs, AZ | 34.01 | -111.99 | 190.98 | |
| 09510200 | Sycamore Creek near Fort McDowell, AZ | 33.81 | -111.47 | 428.2 | |